\documentclass[journal, 12pt, draftclsnofoot, onecolumn]{IEEEtran}
\ifCLASSINFOpdf
\else
\fi

\usepackage{bm}
\usepackage{CJK}
\usepackage{indentfirst}
\usepackage{amsmath}
\usepackage{multirow}
\usepackage{threeparttable}
\usepackage{graphicx}
\usepackage{subfigure}
\usepackage{color}
\usepackage{booktabs}
\usepackage{amsmath,amssymb}
\usepackage{epstopdf}
\usepackage{inputenc}
\usepackage{diagbox}
\usepackage{cite}
\makeatletter
\newif\if@restonecol
\makeatother

\usepackage[linesnumbered,ruled,vlined]{algorithm2e}   
\usepackage{algpseudocode}
\usepackage{amsmath}

\begin{document}

\title{Time-Series Regeneration with Convolutional Recurrent Generative Adversarial Network for Remaining Useful Life Estimation}


\author{Xuewen~Zhang,~\IEEEmembership{}
        Yan~Qin,~\IEEEmembership{}
        Chau~Yuen,~\IEEEmembership{Fellow~IEEE,}
        Lahiru~Jayasinghe,~\IEEEmembership{}
        and Xiang~Liu~\IEEEmembership{}
\thanks{This work was supported by the A*STAR-NTU-SUTD Joint Research Grant on Artificial Intelligence Partnership under Grant RGANS1906 and in part supported by the National Natural Science Foundation of China under Grant 61903327. (Corresponding author: Yan Qin)}
\thanks{X.W. Zhang and X. Liu are with the Department of Software and Microelectronics, Peking University, Peking, 100871 China. ( e-mail: zhangxuewen2018@qq.com, xliu@ss.pku.edu.cn).}
\thanks{Y. Qin, C. Yuen, and L. Jayasinghe are with the Engineering Product Development Pillar of Singapore University of Technology and Design, 8 Somapah Road, 487372 Singapore. (e-mail: yan$\_$qin@sutd.edu.sg, yuenchau@sutd.edu.sg, lahiruaruna@gmail.com)}
}


%

\markboth{}
{Shell \MakeLowercase{\textit{et al.}}: Bare Demo of IEEEtran.cls for IEEE Journals}
%



\maketitle
\renewcommand{\baselinestretch}{2.0}
\begin{abstract}
For health prognostic task, ever-increasing efforts have been focused on machine learning-based methods, which are capable of yielding accurate remaining useful life (RUL) estimation for industrial equipment or components without exploring the degradation mechanism. A prerequisite ensuring the success of these methods depends on a wealth of run-to-failure data, however, run-to-failure data may be insufficient in practice. That is, conducting a substantial amount of destructive experiments not only is high costs, but also may cause catastrophic consequences. Out of this consideration, an enhanced RUL framework focusing on data self-generation is put forward for both non-cyclic and cyclic degradation patterns for the first time. It is designed to enrich data from a data-driven way, generating realistic-like time-series to enhance current RUL methods. First, high-quality data generation is ensured through the proposed convolutional recurrent generative adversarial network (CR-GAN), which adopts a two-channel fusion convolutional recurrent neural network. Next, a hierarchical framework is proposed to combine generated data into current RUL estimation methods. Finally, the efficacy of the proposed method is verified through both non-cyclic and cyclic degradation systems. With the enhanced RUL framework, an aero-engine system following non-cyclic degradation has been tested using three typical RUL models. State-of-art RUL estimation results are achieved by enhancing capsule network with generated time-series. Specifically, estimation errors evaluated by the index score function have been reduced by 21.77$\%$, and 32.67$\%$ for the two employed operating conditions, respectively. Besides, the estimation error is reduced to zero for the Lithium-ion battery system, which presents cyclic degradation.
\end{abstract}


\IEEEpeerreviewmaketitle

\section{Introduction}
\IEEEPARstart{R}{emaining} useful life (RUL) estimation plays a vital role in condition-based maintenance, which estimates the useful life left on diversely daily used components and industrial systems in advance before they thoroughly break down \cite{Ref1}. RUL estimation enables early warning for a maintainer to avoid unplanned downtime of equipment that may cause catastrophic consequences \cite{Ref2}. For instance, estimation of remaining hours for running aero-engines contributes to the maintenance plan and avoids accidents; estimation of available discharging cycles for Lithium-ion batteries (LiB) powered devices helps to timely replace failed batteries. With the ever-increasing strict requirements of operation safety, providing accurate RUL estimation becomes a significant concern in both academic and industrial societies.

Nowadays, advanced machine learning methods, especially the convolutional neural network (CNN) \cite{Ref3} and the recurrent neural network \cite{Ref4}, have been the mainstream for RUL estimation. As a challenging issue in many industrial scenarios, like machinery, aerospace engineering, energy storage, etc., RUL has already attracted substantial attention in each research field. According to the data presentation of failure components, the current RUL methods could be broadly classified into two categories regardless of associated disciplines, in which one is the non-cyclic degradation pattern, and the other one is cyclic degradation. Non-cyclic degradation refers to that a particular object continuously degrades over time, which is one-time. Unlike the non-cyclic degradation pattern, cyclic degradation follows an inconsistently repetitive but similar degradation behavior for the same object, leading to performance indicator deterioration from cycle to cycle. For RUL estimation of non-cyclic degradation pattern, Zheng et al. \cite{Ref5} applied deep long short-term memory network (LSTM) into RUL estimation to learn the long-term degradation trend of faulty components. Experimental results supported to conclude that LSTM outperformed other traditional methods, e.g., support vector machine, with respect to RUL estimation. Wu et al. \cite{Ref6} proposed a similar RUL estimation structure to \cite{Ref5}, but differential features of measurements were additionally fed to present extra degradation information. Jayasinghe et al. \cite{Ref7} incorporated temporal convolutions into LSTM to comprehensively consider both long-term and short-term time dependencies. Assigning weights for different layers, Chen et al. \cite{Ref8} put forward an attention LSTM-based RUL estimation model for machinery process by jointly considering features at each layer rather than the final features. Recently, the capsule network (CapsNet) has been applied for RUL estimation, which has the ability to present features in high-dimension space \cite{Ref9}. For RUL estimation of cyclic-degradation pattern, the mainstream machine learning algorithms have been reported. Taking the LiB system as an example, informative measurements, like charging/discharging voltage, charging/discharging current, and impedance, can be used for the RUL estimation by learning the regression relationship between these features and the degradation performance [10], [11]. In additional to the regression way to perform RUL estimation, it is feasible to treate the RUL estimation in an autoregression way due to the cyclic degradation nature through learning past performance to forecast the future ones  [12], [13]. GAN is capable of forecasting by constructing the generator as an autoregressive model. For instance, GAN is employed for stock price prediction \cite{Ref19}, the typhoon trajectory \cite{Ref20}, and failure frame \cite{Ref21}.

Despite the employment of various advanced deep learning algorithms, the essence of these models [3]-[16] lies in regressing the concerned performance indicator, which may be high-cost or time-consuming to measure, on available sensor data. In other words, a regression relationship will be developed for two groups of specific variables. The methods mentioned above inferred faulty evolution patterns from measured time-series, assisting preventive maintenance and quantitative indication. Therefore, the acquisition of time-series data is the primary but fundamental step to obtain satisfied estimation. This step involves a large amount of data to train a good neural network to avoid overfitting and keep generalization ability. In the above literature, the data are run-to-failure time-series that record long-term degradation procedure of machines or components from health status to failure, indicating a whole time-series as a sample. Although it is not difficult to collect plenty of data in the era of Big Data, collecting enough run-to-failure time-series is still an obstacle in practice. Taking the turbofan engine as an example, it may take months even years for a turbofan engine to naturally experience initial degradation until it completely steps into failure status. Practically, failure status is undesirable and prohibited to avoid safety violation and catastrophic damage. Hence parts are replaced much before failure due to safety reasons. As a result, data shortage in terms of failure cycles is an ever-increasing challenge for advanced machine learning-based RUL estimation to successfully fulfill the task of health prognosis.

As an alternative to physical experiments, data generation technology is a promising solution in enriching data, creating realistically synthetic data through learning data distribution from real sequences. Among data generation algorithms, the most important one is the generative adversarial network (GAN) proposed in 2014 \cite{Ref14}. It succeeds in introducing a novel adversarial modelling framework, which includes a generative network and a discriminative network. These two networks construct a two-player minimax game, using CNNs to sidestep the calculation of intractable likelihood functions. Since GAN has been proposed, its superiorities in data generation have been noticed and attracted extensive studies in image synthesis and natural language processing. Recently, the success of GAN in synthesizing sequences raises attention \cite{Ref15}. Li et al. \cite{Ref16} applied GAN to increase available tokens for discrete sequences. Ramponi et al. \cite{Ref17} proposed a time-conditional GAN for sequence augmentation, which introduced a deconvolutional neural network in generative step and CNN in the discriminative network. Since the above-mentioned methods [18]-[20] are not designed for time-series generation, the characteristics of temporal correlations between past samples and future data, which is the nature of industrial time-series, cannot be ensured if they are directly applied. Besides, industrial time-series are always multivariate rather than a single variable, presenting correlations between variables. To solve these problems, Esteban et al. \cite{Ref18} attempted to generate medical time-series by replacing CNN with LSTM in the original GAN framework to capture long-term temporal correlation. However, this method may not be suitable for run-to-failure data, in which inherent complexity and strong dynamics are observed. That is, multivariate in degradation procedure simultaneously present serious non-stationarity and long-term temporal correlation within each cycle. In addition, some industrial time-series are cyclic in nature, e.g., the battery system that is charged and discharged over repeating cycles. These characteristics cause challenges to generate high-quality time-series using current methods. Moreover, to the best of the authors' knowledge, few studies have been reported about the hybrid RUL model with time-series regeneration, which is a non-destructive and safe way to overcome the data shortage problem of RUL estimation.

To address the aforementioned problems, an enhanced RUL estimation framework is proposed, which combines the proposed time-series generation method, named convolutional recurrent GAN (CR-GAN), with current RUL estimation models. First, both generator and discriminator are improved in the proposed CR-GAN. In the generator, a two-channel fusion model is given to simultaneously learn both spatial and temporal evolution nature of the concerned multivariate time-series. To achieve this, CNN and LSTM are combined to ensure learning ability, and initial multivariate time-series are generated according to the hidden relationship with consideration of external information. Next, the discriminator employs LSTM to verify the performance of the generated data, acting as an evaluator. The interactions between the generator and the discriminator ensure that the generated data are realistic (i.e., similar to the real data). Finally, with augmented time-series, the proposed RUL framework is tested and compared in actual industrial scenarios. The contributions of the proposed method are summarized below:

1) An enhanced RUL estimation concept is first proposed with the idea of time-series self-generation to overcome the shortage of run-to-failure data in a low-cost, rapid, and safe way, improving the current RUL estimation models with more data;

2) CR-GAN model is designed to ensure the quality of generated time-series, serving both non-cyclic and cyclic degradation patterns in a unified framework;

3) We evaluate our proposed method with an aero-engine system \cite{Ref22} and a LiB battery system \cite{Ref23} both from NASA to demonstrate its effectiveness in reducing RUL estimation error with generated time-series.

The remainder of this article is structured as follows: In Section II, the RUL estimation problem is classified according to different degradation patterns, following with the motivation. The proposed method is detailed in Section III, including data preparation, specific steps of the enhanced RUL estimation framework, and online implementation. Section IV illustrates the efficacy of the proposed method in typical systems. Conclusions are drawn in the last section.

\begin{figure*}[t]
\centering
\includegraphics[scale=0.85]{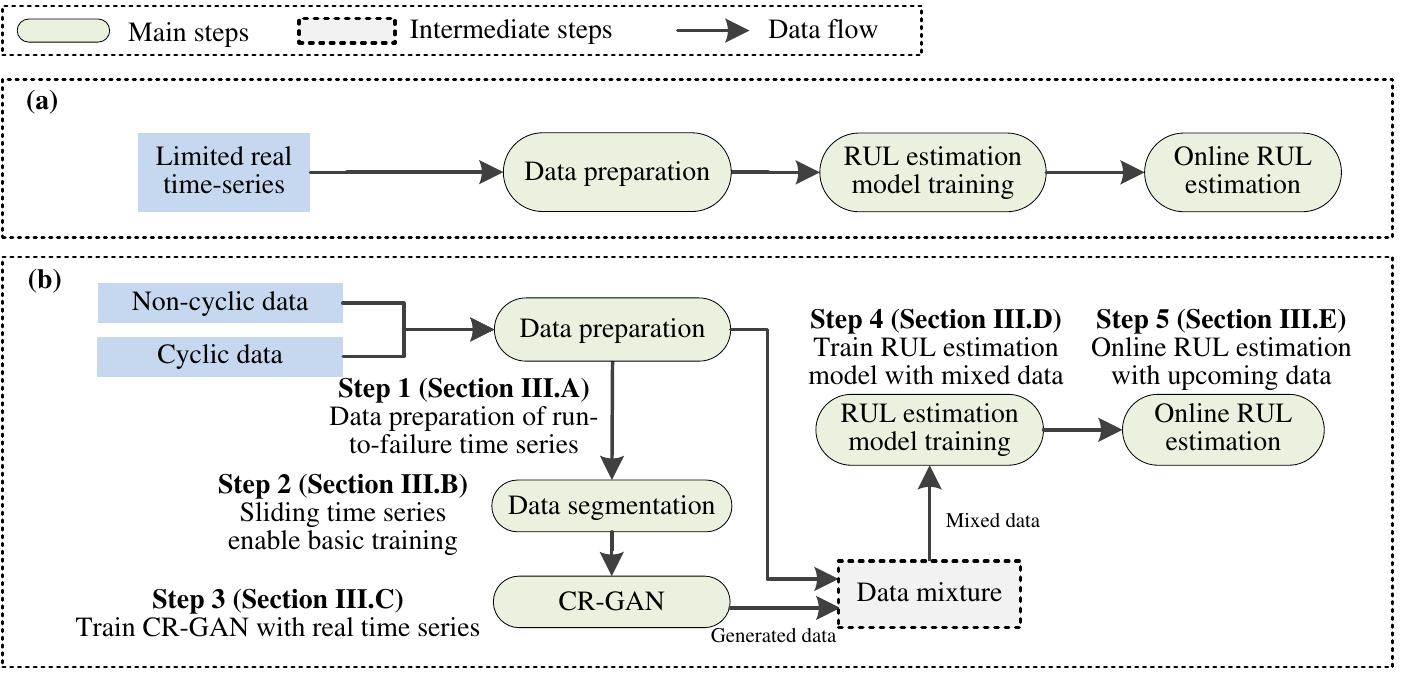}
\caption{The main components and flowchart of (a) the traditional methods and (b) the proposed method.}
\label{MyFig1}
\end{figure*}

\section{Problem Formulation and Motivation}
This section summarizes current RUL estimation problems into two groups according to different degradation patterns, including non-cyclic degradation and cyclic degradation. After that, the motivation for the proposed method is presented.
\subsection{Problem Formulation of RUL Estimation}
\subsubsection{RUL estimation of non-cyclic degradation pattern}
Non-cyclic degradation refers to that a certain object continuously degrades over time, which is one-time. For example, when an aero-engine begins to fail, its RUL decreases until it totally breaks down. Thereby, the non-cyclic degradation of an object constructs a two-dimensional data matrix. Assuming $J$ process variables are measured at discrete time instance $k=1,2,...,K_i$, a matrix $\textbf X_i (J \times K_i)$ is formed, where $K_i$ is the length of the $i^{th}$ object. Correspondingly, a vector recording the remaining life is $\textbf Y_i=[K_i, K_i-1, ..., 1]$, which decreases linearly with the increase of time. Given there are $I$ run-to-failure objects, a series of pair-wise data matrixes  $\{\textbf X_1, \textbf Y_1\}$,$\{\textbf X_2, \textbf Y_2\}$,...,$\{\textbf X_I, \textbf Y_I\}$ are prepared, each of which corresponds to a certain object. For RUL estimation, the goal is to find a degradation pattern ${\phi_{nc}}$ from objects 1, 2, $..., I$ by minimizing the estimation error as follows,
\begin{equation}
\min \limits_{\phi_{nc}} \sum_{i=1}^{I} \sum_{k=1}^{K_i} \|\textbf Y_i(k)- {\phi_{nc}}(\textbf X_i(1:k))\|_2^2
\label{Eq1}
\end{equation}
where $\textbf X_i(1:k)$ is time-series ranged from time 1 to $k$, and $\textbf Y_i(k)$ is the $k^{th}$ sample in $\textbf Y_i$.

\subsubsection{RUL estimation of cyclic degradation pattern}
Different from the non-cyclic degradation pattern, cyclic degradation follows an inconsistently repetitive but similar degradation behavior for the same object, leading to performance indicator deterioration from cycle to cycle. For instance, after a discharging cycle, LiB's real capacity may slightly drop. After thousands of charging and discharging cycles, the capacity loss will cumulate, and its real capacity will reach the end of life (EOL) threshold, indicating the replacement of LiB. For such a kind of degradation pattern, assuming $J$ process variables are measured at discrete time instance $k=1,2,...,K_c$ in the $c^{th}$ cycle, a data matrix $\textbf X_c (J \times K_c)$ is formed, where $K_c$ is the cycle length. Correspondingly, the performance indicator is recorded as $Y_c$, which presents a cumulative influence of the whole cycle. By collecting $C$ run-to-failure cycles, a series of pair-wise cycling data $\{\textbf X_1, Y_1\}$, $\{\textbf X_2, Y_2\}$, ..., $\{\textbf X_C, Y_C\}$, each of which corresponds to a certain cycle. For RUL estimation, the goal is to find a relationship ${\phi_c}$ from cycles 1, 2, $..., C$ by minimizing the estimation error below,
\begin{equation}
\min \limits_{\phi} \sum_{c=1}^{C} \|Y_c- {\phi_c}(\textbf X_c)\|_2^2
\label{Eq2}
\end{equation}

Due to cyclic degradation nature, it is promising to consider RUL estimation in an autoregression modeling way through directly learning past performance to forecast the future ones. However, this way may have a real-time requirement on performance indicators. Since this article mainly focuses on the regression way through easy-measured sensor data to predict the concerning performance, the autoregressive-based RUL estimation models are not included in this classification.

\subsection{Motivation}
From the above analysis, data matrixes of both degradation patterns can be arranged as a three-dimensional matrix $\textbf X(N \times J \times K_n)$, where the symbol $N$ indicates the number of components $I$ in non-cyclic degradation or cycles $C$ for the same object in cyclic degradation, and $K_n$ is the corresponding sample length. For readability, $\textbf X(N \times J \times K_n)$ is used as a general form in the following analysis.

Strong dynamics and non-stationary characteristics are observed for both types of degradation patterns, resulting in varying data distribution of process variables. Although the number of variables $J$ is fixed, it is more reasonable to consider the variable dimension combining sampling time. That is, the original data matrix with dimensions $J \times K$ is extended to $1 \times JK$ by treating a complete run-to-failure time-series as a basic data unit. Usually, $JK$ is greatly large than $N$, resulting in a shortage of training cycles. Instead of increasing the amount of cycling data physically, it is more affordable to enrich data through a data-driven way. If we could model the degradation nature, it is possible to construct a realistic simulation to mimic it. The above analysis is the basic idea of the proposed method, which meets well with the purpose of GAN.

\begin{figure}[tb]
\centering
\includegraphics[scale=1]{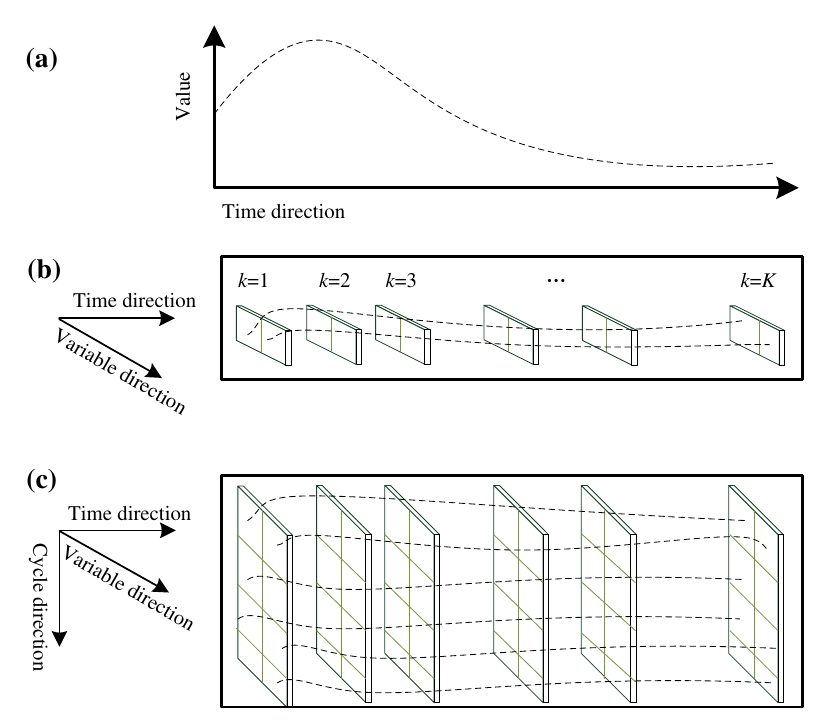}
\caption{The data presentation of (a) a single variable, (b) non-cyclic degradation pattern, and (c) cyclic degradation pattern.}
\label{MyFig2}
\end{figure}

\section{RUL Estimation with Time-series Generation}
In this section, the organization and highlights of the proposed method are illustrated in Fig. \ref{MyFig1}, in which five sequential parts are given. First, data preparations are conducted on both non-cyclic and cyclic degradation systems to transform run-to-failure cycles into a three-dimensional data matrix without changing their physical structure. Next, data segmentation is performed to increase the data amount. After that, the inherent evolution trend of degradation patterns is learned through the proposed CR-GAN, enabling the generation of similar degradation trajectories with the same nature. Then generated temporal sequences combined with the real run-to-failure data sequences are mixed as new inputs to train RUL estimation models, improving its estimation ability. At last, when a new time-series is available, the well-trained RUL estimation model can be adopted for online prediction.

\subsection{Data Preparation}
The degradation system always consists of dozens of slowly varying variables, as shown in Fig. \ref{MyFig2}(a). As mentioned in Section II.A, the data structure in a complete run-to-failure object following a non-cyclic degradation pattern is denoted as $\textbf X_i(J \times K_i)$ as shown in Fig. \ref{MyFig2}(b). A three-dimensional form can be arranged by putting data matrixes together without considering the order of objects. However, for cyclic degradation, cycling data should be arranged according to the ascending order of cycle number. Therefore, a series of sequential two-dimension data matrixes are extended to a three-dimensional data matrix $\textbf X$ as shown Fig. \ref{MyFig2}(c), which takes spatial correlation over cycles into account inherently.

Denote the maximum and the minimum of each variable among the data matrix $\textbf X$ as $\textbf M_{max}=[M_1,M_2,...,M_J]$ and $\textbf V_{min}=[V_1,V_2,...,V_J]$, respectively. Each sample is scaled using minimax normalization into range [0, 1] as follows,
\begin{equation}
\begin{aligned}
{\overline x}(n,j,k)= \frac {x(n,j,k) - V_j} {M_j}
\end{aligned}
\label{Eq6}
\end{equation}
where $x(n,j,k)$ is the $j^{th}$ variable in the $k^{th}$ time index of the $n^{th}$ cycle, $V_j$ is the minimum of the $j^{th}$ variable, $M_j$ is the maximum of the $j^{th}$ variable, and $\overline x(n,j,k)$ is the normalized sample in data matrix $\overline {\textbf X}$.

\subsection{Data Segmentation}
To enable the basic training of the CR-GAN model, data segmentation is performed after normalization. The $n^{th}$ complete run-to-failure time-series $\mathbf X_n$ in $\overline {\textbf X}$ is divided into a series of overlapped local trajectories. Each local trajectory is obtained by sliding the complete time-series along the time direction with a fixed window size $L$. A complete time-series is transformed into many locals trajectories to enrich the training data in a step-wise way. Taking engine data \cite{Ref22} as an example, Fig. 3 illustrates the procedure of data segmentation with step size 1 using the first two variables. By performing the data segmentation for all run-to-failure time-series, the number of $N$ complete time-series is enlarged to $N(K-L+1)$, where $K$ is the total length of a run-to-failure time-series. For brevity, data matrixes after segmentation are still denoted as $\overline {\textbf X}$. As a tunable parameter, the value of $L$ is suggested to choose 80$\%\sim$90$\%$ of $K$, enriching the amounts of time-series from 10 to 20 times without changing of the degradation behavior.

\begin{figure}[t]
\centering
\includegraphics[scale=0.36]{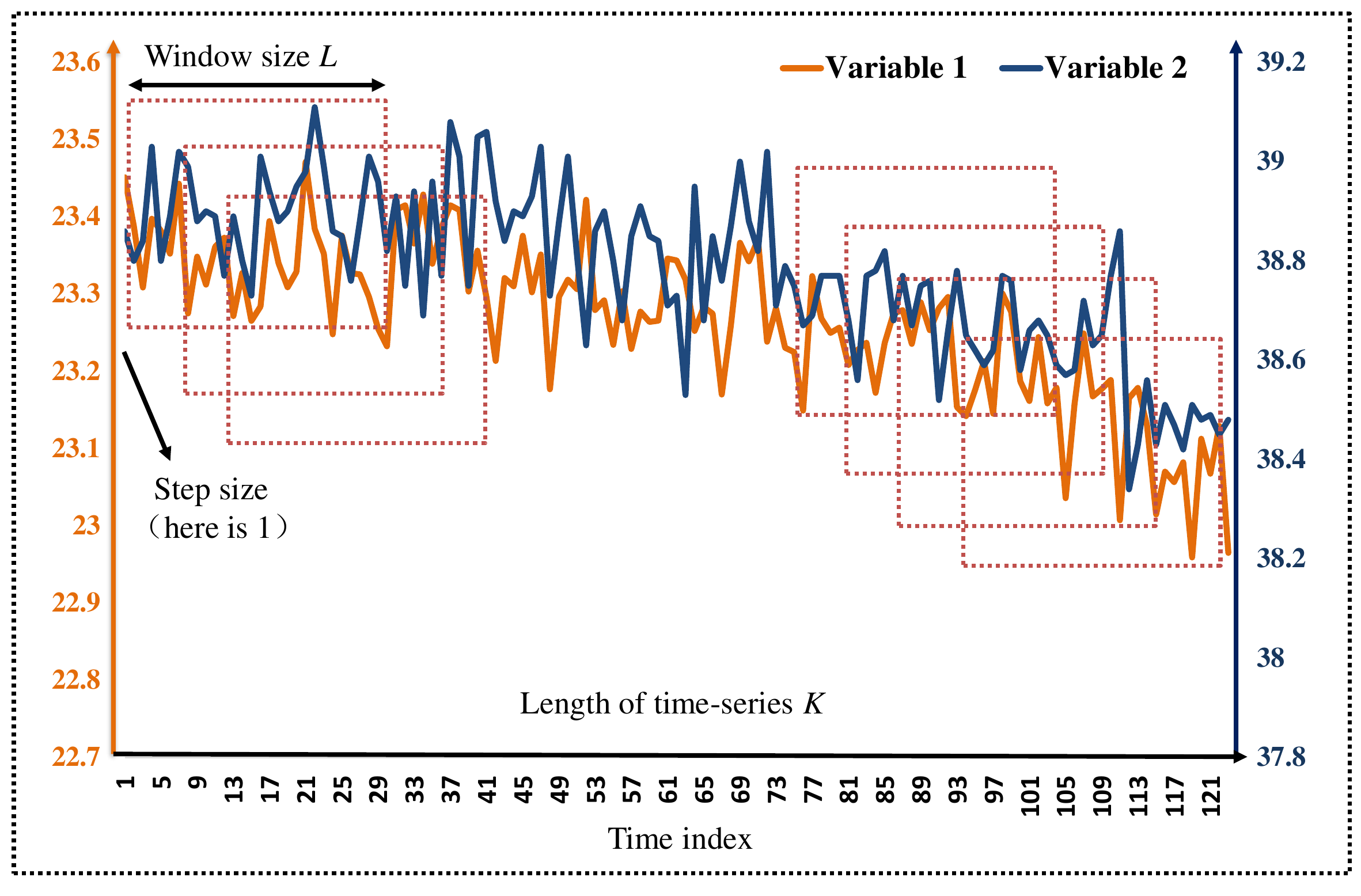}\\
\begin{center}
\caption{The illustration of data segmentation for a complete run-to-failure time-series from engine data.}
\end{center}
\label{MyFig3}
\end{figure}

\subsection{The CR-GAN Model}
Two kinds of characteristics need to be carefully analyzed in generating high-quality run-to-failure time-series: the first one is temporal correlations between the past samples and future data in each cycle, and the second one is the similarities and dissimilarities between different cycles along cycle direction. These characteristics are ensured by a novel Convolutional Recurrent GAN (CR-GAN) designed with the combination of LSTM and CNN.

\begin{figure*}[t]
\centering
\subfigure[The detailed structure of generative network]{
\begin{minipage}[a]{0.8\textwidth}
\includegraphics[width=1\textwidth]{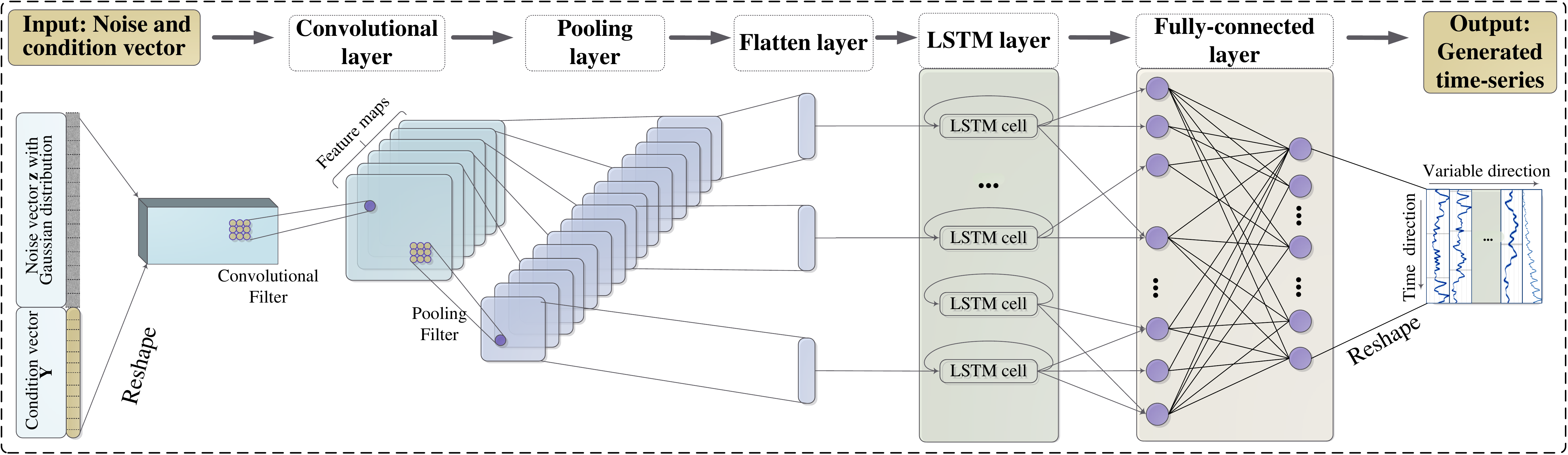} \\
\end{minipage}}
\subfigure[The detailed structure of discriminative network]{
\begin{minipage}[a]{0.8\textwidth}
\includegraphics[width=1\textwidth]{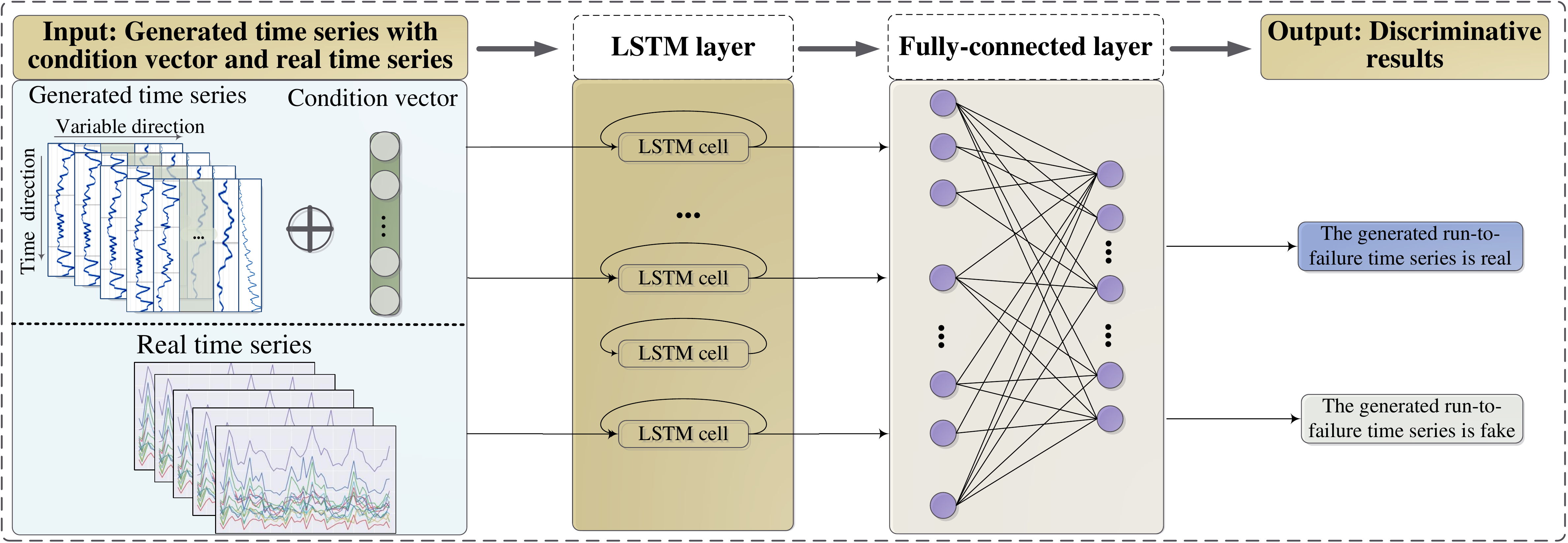} \\
\end{minipage}}
\caption{The detailed network structure for (a) Generator and (b) Discriminator.}
\end{figure*}
\label{MyFig4}

\subsubsection{The structure of CR-GAN Model}
The generative model $G$ introduces both convolutional ability and long-term prediction ability to generate run-to-failure time-series. As shown in Fig. 4(a), the generator integrates convolutional layers and LSTM layers together. Specifically, CNN layers focus on describing spatial features of the sequence, and LSTM layers aim at learning temporal correlations between the past samples and the future samples. Moreover, the fusion of spatial and temporal information is achieved by putting these two networks into an integrated  network. Here, the noise vector is assumed to follow Gaussian distribution, which has been widely adopted in most literature [14]-[21], [24]-[27]. Through the temporal learning ability, the generator outputs a series of generated time-series for further evaluation of the discriminator. Fig. 4(b) describes the network structure of the discriminator, where the quality of generated time-series is evaluated by classifying generated time-series and real time-series. The discriminator is implemented using the LSTM network and the fully-connected layer. For concisely, only one convolutional layer or one LSTM layer have been presented in both the generator and the discriminator, and more layers can be easily stacked to achieve better performance during practical training.

For non-cyclic data, data distributions of variables are varying along time direction for the same object. However, for cyclic data, the data distribution of a variable in a specific cycle is treated as a data unit, and it slowly changes over cycles. Thereby, the time index of non-cyclic data and cycle index in cyclic data could serve as an auxiliary indicator to achieve better RUL estimation. That is, the time index can be served as the condition vector $\mathbf Y$.

By introducing condition information $\mathbf Y$ into Generator $G$ and Discriminator $D$, CR-GAN replaces $D(\textbf x)$ and $G(\textbf z)$ with $D(\textbf x | \mathbf Y)$ and $G(\textbf z | \mathbf Y)$, respectively. At this point, both the generator and the discriminator take extra information $\textbf Y$ as a condition, which is fed into the two models as additional input as shown in Figs. 4(a) and 4(b). Through the continual interaction between Discriminator and Generator, unqualified time-series will be filtered out via iterative solution, which is specified in the following part.

\subsubsection{Iterative solution of CR-GAN Model}
The loss of Discriminator consists of two parts. The first one is the probability of generated samples belonging to the real data, which is as large as possible. The second one is the probability of generated samples belonging to the synthetic data, which is as small as possible. As a result, the objective function of Discriminator is constructed as follows,
\begin{equation}
\begin{aligned}
max V_D & = \mathbb{E}_{\textbf{x} \sim p_{\text {data}}(\textbf {x})}[\log D(\textbf x| \mathbf Y)] \\
& + \mathbb{E}_{\textbf{z} \sim p_{\textbf{z}}(\textbf{z})}[\log (1-D(G(\textbf z| \mathbf Y)))
\end{aligned}
\end{equation}
where $p_z(\textbf z)$ is the aprior distribution of noise variables $\textbf z$, $G(\textbf z)$ is the generated data using Generator $G$, $D(\textbf x)$ presents the probability that $\textbf x$ came from the given real data, and $D(G(\textbf z))$ stands for the probability that generated data come from the distribution of the given real data.

\begin{algorithm}[htb]
\caption{Iterative solution of CR-GAN}
\LinesNumbered
\KwIn{Normalized real run-to-failure time-series $\overline {\mathbf X}$, noise vector $\mathbf z$ follows Gaussian distribution}
\KwOut{Generator $\mathbf \Theta_g$ and Discriminator $\mathbf \Theta_d$}
Initialize $\mathbf \Theta_g$ of Generator and $\mathbf \Theta_d$ of Discriminator \\
\Repeat
{$\mathbf \Theta_g$ and $\mathbf \Theta_d$ converged}
{\Repeat (//Update the discriminator network)
{Given iterations (here is 3)}
{Sample $m$ noise vector $\mathbf z_1, \mathbf z_2, ..., \mathbf z_m$ from the prior distribution $P_{z}$ and pick up time index ${\mathbf Y}_m$ from known range; \\
Generate $m$ time-series with generative network $\tilde {\mathbf x}_m=G(\mathbf z_m|{\mathbf Y}_m)$; \\
Update discriminator parameter $\mathbf \Theta_d$ to maximize,\\
$V_D = \frac{1}{m} \sum_{i=1}^{m} logD(\mathbf x_i | \mathbf Y_i) + \frac{1}{m} \sum_{i=1}^{m} log(1-D(\tilde {\mathbf x}_i | {\mathbf Y}_i))$\\
$\mathbf \Theta_d \leftarrow \mathbf \Theta_d + \nabla \tilde V_D(\mathbf \Theta_d)$ \\}
\Repeat (//Update the generator network)
{}
{Sample another $m$ noise vector $\mathbf z_1, \mathbf z_2, ..., \mathbf z_m$ from the prior distribution $P_{z}$ and pick up time index ${\mathbf Y}_m$ from known range; \\
Update generator parameter $\mathbf \Theta_g$ to minimize, \\
$V_G = \frac{1}{m} \sum_{i=1}^{m} logD(\mathbf x_i | \mathbf Y_i) + \frac{1}{m} \sum_{i=1}^{m} log(1-D(G(\mathbf z_i| \tilde {\mathbf Y}_i)))$\\
$\mathbf \Theta_g \leftarrow \mathbf \Theta_g + \nabla \tilde V_G(\mathbf \Theta_g)$ \\}
}
\end{algorithm}

The discriminator is trained according to the given number of iterations. Next, the generator loss $V_G$ is minimized by keeping weights of the discriminator constant as follows,
\begin{equation}
min V_G = \mathbb{E}_{\textbf{z} \sim p_{\textbf{z}}(\textbf{z})}[\log (1-D(G(\textbf z)))]
\end{equation}

The training of CR-GANs involves optimizations of $V_D$ and $V_G$ iteratively. Conceptually, the training procedure corresponds to a minimax two-player game as below,
\begin{equation}
\begin{aligned}
\min _{G} \max _{D} V(D, G) & = \mathbb{E}_{\textbf{x} \sim p_{\text {data}}(\textbf {x})}[\log D(\textbf x| \mathbf Y)] \\
& + \mathbb{E}_{\textbf{z} \sim p_{\textbf{z}}(\textbf{z})}[\log (1-D(G(\textbf z| \mathbf Y)))]
\end{aligned}
\label{Eq1}
\end{equation}
where the ideal objective is the discriminator cannot distinguish the generated samples from the training samples.

The solving procedure of CR-GAN is specified in \textbf {Algorithm 1}, in which Discriminator and Generator are iteratively executed until converge. Feeding $\overline {\textbf X}$ into the proposed CR-GAN with additional information $\textbf Y$, a series of qualified generated run-to-failure cycles are generated as ${\textbf X_g}(N_g \times J \times K)$, where $N_g$ is the number of generated cycles.

{\textit{Remark:}} Due to the randomness of the input noise vector, irrational time-series are possibly generated with CR-GAN. The RUL estimation ability may deteriorate if a false degradation pattern is learned. To filter out these irrational time-series, we employ a ``Train on Real, Test on Synthetic (TSTR)'' concept \cite{Ref18} to recognize the unqualified time-series. In this approach, the real run-to-failure time-series are trained as a supervised model using CNN. Then we input the generated time-series using CR-GAN into the trained classification model. The irrational time-series will be identified if it does not belong to the class of real time-series. These irrational time-series will be discarded and not used in subsequential modeling.

\subsection{Enhanced RUL Model with Data Generation}
With the generated data, the enhanced RUL framework is constructed. By merging ${\textbf X_g}$ into $\overline {\textbf X}$ along cycle-direction, a new data matrix ${\textbf X_n}$ is formed with dimensions $N_n\times J \times K$, in which $N_n=N_g + N$. Selecting a proper RUL estimation model, and training the model structure with the matrix ${\textbf X_n}$. Also ${\textbf X_n}$ can be directly fed into existing RUL estimation models with slight adjustments of tunable parameters, such as the number of nodes in LSTM or CNN layers. The RUL estimation relationship ${\phi_{nc} (\cdot)}$ or ${\phi_{c} (\cdot)}$ as shown in Eqs. 1 and 2 are achieved when the iteration epoches are reached or the iteration accuracy are met.

\subsection{Online Estimation using the Enhanced RUL Model}
If a new data vector $\textbf x_{new}(J \times 1)$ from a non-cyclic component or a new cycling data matrix $\textbf X_{new}(J \times K_{new})$ from cyclic component is upcoming, all variables are first normalized into the range $[0,1]$ with the normalization information obtained in Eq. 3, in which $K_{new}$ is the cycle length of the new cycle. For clarity, the normalized data information is still denoted as $\textbf x_{new}$ and $\textbf X_{new}$, respectively.

With well-trained model ${\phi_{nc} (\cdot)}$ and ${\phi_{c} (\cdot)}$, the estimated RUL is calculated by inputting $\textbf x_{new}$ and $\textbf X_{new}$ below,
\begin{equation}
RUL_{new}=\left\{\begin{array}{ll}
{\phi_{nc}}(\textbf x_{{new}}) & \text{Non-cyclic degradation} \\ {\phi_{c}}(\textbf X_{{new}}) & \text{Cyclic degradation}
\end{array}\right.
\label{Eq8}
\end{equation}
where $RUL_{new}$ is the remaining times of non-cyclic component or the remaining cycles of a cyclic component.

To measure performance of the enhanced RUL estimation model, the indices scoring function ($SF$) \cite{Ref5} and root mean square error ($RMSE$) \cite{Ref28} \cite{Ref29} are used. $RMSE$ equally weights both early estimation and delay estimation, which is defined as follows,
\begin{equation}
RMSE=\sqrt{\frac{1} {N_t} \sum_{c=1}^{N_t}(d_c)^2}
\end{equation}
where $d_c$ is the difference between $RUL_{new}$ and the real useful life left, and $N_t$ is the number of components in non-cyclic degradation or testing cycles in cyclic degradation.

Specifically, index $SF$ is a piece-wise function calculated as follows,
\begin{equation}
SF=\left\{\begin{array}{ll}
\sum_{c=1}^{N_t} {(e^{-\frac{d_{c}}{13}}-1)} & {d_{c}<0} \\ \sum_{c=1}^{N_t} {(e^{\frac{d_{c}}{10}}-1)} & {d_{c} \geq 0}
\end{array}\right.
\label{Eq8}
\end{equation}

Index $SF$ gives a large punishment for the case when lagged estimation is observed. With these two metrics, performance of the enhanced RUL estimation method can be quantitatively compared with the results without generated data. Both values of $SF$ and $RMSE$ are the smaller the better. Considering $RMSE$ and $SF$ are overall metrics by averaging or summing all weighted estimation errors, it is recommended to further analyze the distribution of estimation errors. For two methods with a similar value of $RMSE$, the method who has more estimation errors distributed in the low-error zones will be preferred.

\section{Results and discussions}
To demonstrate the efficacy of the proposed framework, three cases are employed in this section, one of which is a numerical example and the other two are practical ones. The numerical case focuses on verifying the quality of the generated time-series from an intuitive view. The other two practical cases comprehensively evaluate the proposed framework by simultaneously considering the quality of generated time-series and RUL estimation accuracy from non-cyclic degradation using aero-engine open dataset \cite{Ref22} and cyclic degradation with LiB open dataset \cite{Ref23}.

For fair comparisons, all simulations are conducted on a desktop equipped with 6 processors of Intel Xeon E5-2620 v4 and a multi-graphics processor unit of NVIDIA GeForce GTX 1080Ti. Programming language is Python 3.6 with scientific computing library ``Pandas (0.25.3)'', ``Numpy (1.16.4)'', ``Scipy (1.2.1)'', ``Matplotlib (3.0.3)'', ``Scikit-learn (0.20.3)'', ``Tensorflow (1.15.0)'', and ``Keras (2.3.1)''.
\subsection{Time-series Generation with A Numerical Case}
Two kinds of simulation signals are constructed as real time-series, including a sinusoidal wave and a smooth signal. The $i^{th}$ generated sinusoidal time-series $S_i(k)$ is generated with varying amplitude and frequency, which is given below,
\begin{equation}
{S_i(k) = A_rsin(\omega_i k+ \varphi_i)}
\end{equation}
where $\omega_i$ is frequency randomly selected from range [1, 5], $A_i$ is amplitude randomly picked up from range [0.1, 0.9], and $\varphi_i$ is phase randomly given between [$-\pi,\pi$].

A more complex smooth signal is constructed with time-varying variations. Repeating the process ten times with the same length of 16, and then the corresponding cycling time-series are obtained with dimensions of $10 \times 1 \times 16$. Fig. 5 plots the real waves of the generated time-series using the proposed CR-GAN for the sinusoidal and the smooth signal, respectively. In this case, the quality of the generated sine waves is examined by visual inspection, which is almost the same as given real waves.

\begin{figure}
  \centering
  \includegraphics[scale=0.25]{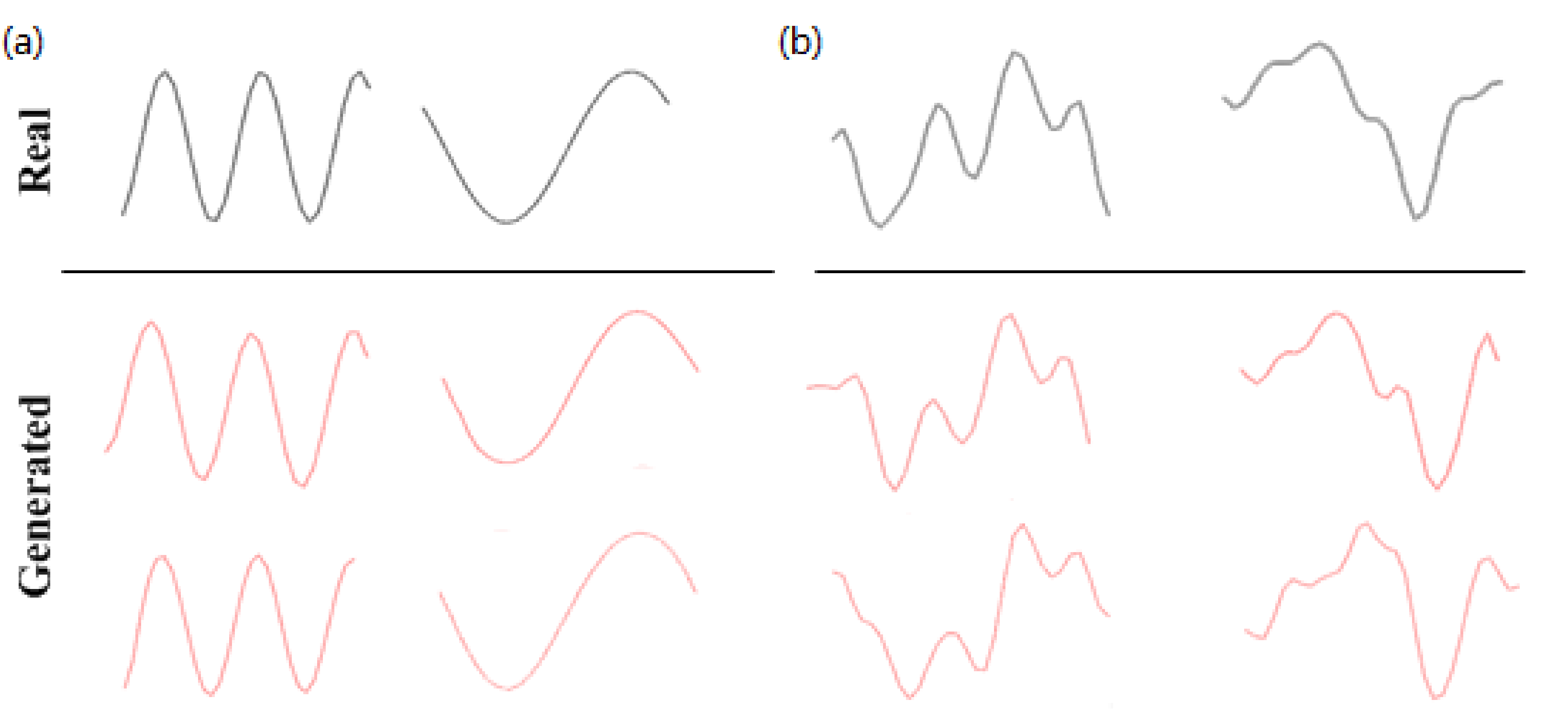}\\
  \caption{Example of real curves and generated curves for (a) sinusoidal signal and (b) smooth signal.}
  \label{MyFig5}
\end{figure}

\begin{figure*}[htb]
\centering
\subfigure[]{
\begin{minipage}[t]{0.5\linewidth}
\centering
\includegraphics[width=8.5cm]{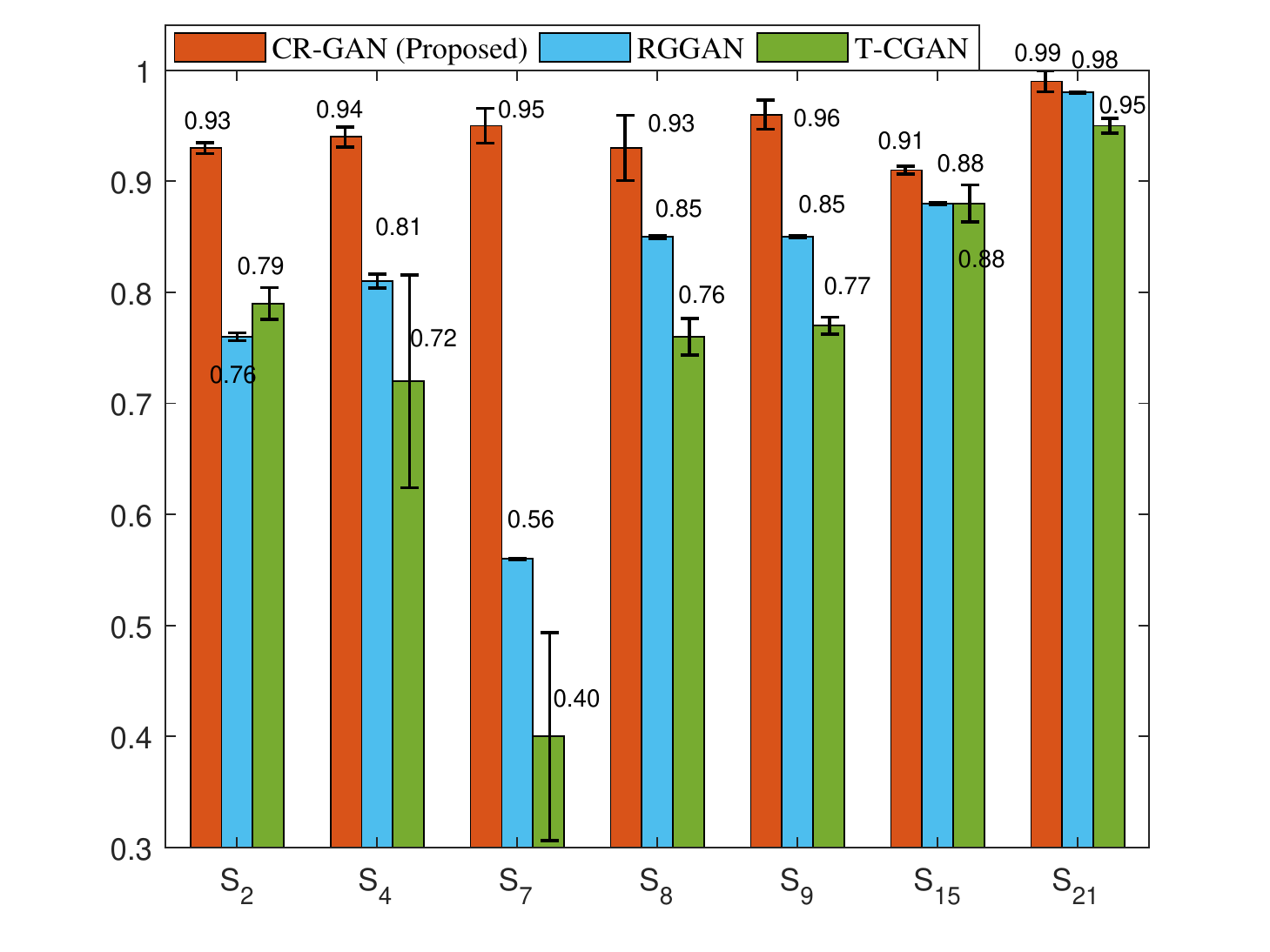}
\end{minipage}}%
\subfigure[]{
\begin{minipage}[t]{0.5\linewidth}
\centering
\includegraphics[width=8.5cm]{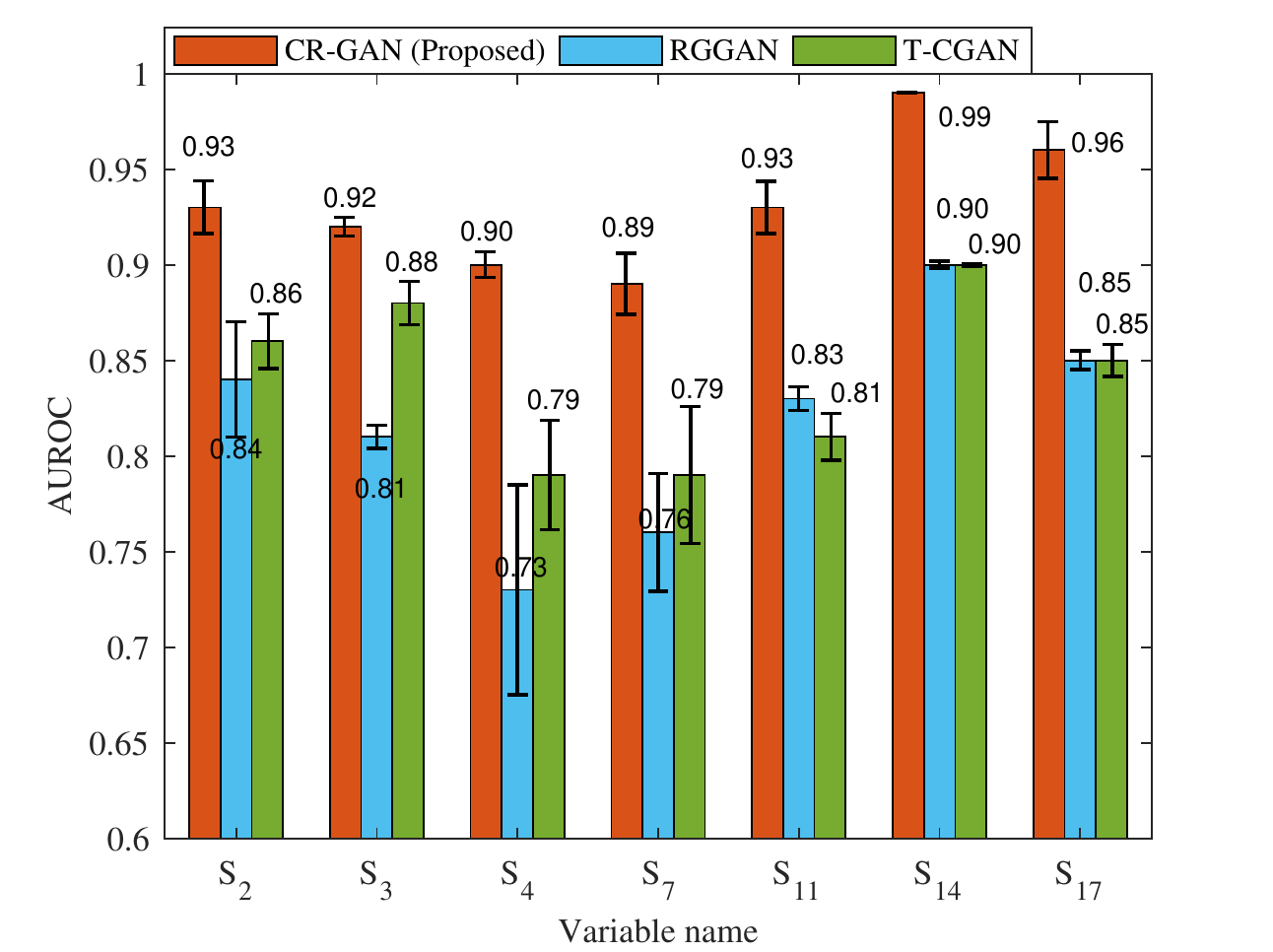}
\end{minipage}}
\centering
\caption{Performance comparisons between the proposed method and its counterparts for generated variables with (a) Dataset FD001 and (b) Dataset FD003.}
\label{MyFig6}
\end{figure*}

\begin{table}[!htb]
\scriptsize
\setlength{\abovecaptionskip}{0.cm}
\setlength{\belowcaptionskip}{-1cm}
\renewcommand{\arraystretch}{1.2}
\caption{Description of variables collected in engine system.}
\label{Table1}
\begin{center}
\begin{threeparttable}
\begin{tabular}{c c c c}
\hline
\toprule
\textbf{No.} & \textbf{Symbol} & \textbf{Variable description} & \textbf{Unit} \\
\hline
1 & $S_1$ & Total temperature at fan inlet & $^o$C \\
\hline
2 & $S_2$ & Total temperature at low pressure compressor outlet & $^o$C \\
\hline
3 & $S_3$ & Total temperature at high pressure compressor outlet & $^o$C \\
\hline
4 & $S_4$ & Total temperature at low pressure turbine outlet & $^o$C \\
\hline
5 & $S_5$ & Pressure at fan inlet & Pa \\
\hline
6 & $S_6$ & Total pressure in bypass-duct & Pa \\
\hline
7 & $S_7$ & Total pressure at high pressure compressor outlet & Pa \\
\hline
8 & $S_8$ & Physical fan speed & r/min \\
\hline
9 & $S_9$ & Physical core speed & r/min \\
\hline
10 & $S_{10}$ & Engine pressure ratio  & 1 \\
\hline
11 & $S_{11}$ & Static pressure at high pressure compressor outlet & Pa \\
\hline
12 & $S_{12}$ & Ratio of fuel flow to Ps30 & 1 \\
\hline
13 & $S_{13}$ & Corrected fan speed & r/min \\
\hline
14 & $S_{14}$ & Corrected core speed & r/min \\
\hline
15 & $S_{15}$ & Bypass ratio  & 1 \\
\hline
16 & $S_{16}$ & Burner fuel-air ratio  & 1 \\
\hline
17 & $S_{17}$ & Bleed enthalpy  & 1 \\
\hline
18 & $S_{18}$ & Demanded fan speed & r/min \\
\hline
19 & $S_{19}$ & Demanded corrected fan speed & r/s \\
\hline
20 & $S_{20}$ & High pressure turbine coolant bleed & lbm/s \\
\hline
21 & $S_{21}$ & Low pressure turbine coolant bleed & lbm/s \\
\bottomrule
\end{tabular}
\end{threeparttable}
\end{center}
\end{table}

\subsection{RUL Estimation for Non-cyclic Degradation Pattern}
In this section, a more complex multivariate dataset, i.e., commercial modular aviation propulsion system simulation (C-MAPSS) \cite{Ref22} \cite{Ref30}, provided by NASA is used for further evaluation. C-MAPSS provides complete degradation procedures of the simulated engine system from normal status to faulty status, which is a typical non-cyclic degradation pattern. According to the number of operating conditions, four datasets in C-MAPSS are classified into two groups, one of which (FD001 and FD003) follows single data distribution and the other one is multi-distribution (FD002 and FD004). Since the proposed CR-GAN does not consider the characteristics of multiple distributions, only the datasets.   FD001 and FD003 are chosen for analysis. For both FD001 and FD003, there are 100 training engines and 100 testing engines, each of which contains 21 variables listed in Table I \cite{Ref22}. Besides, real RUL values of training engines are known as a priori information.

\begin{table}[t]
\scriptsize
\setlength{\abovecaptionskip}{0.cm}
\setlength{\belowcaptionskip}{-1cm}
\renewcommand{\arraystretch}{1.2}
\caption{Specific Configuration of Generator in the proposed CR-GAN.}
\label{Table2}
\begin{center}
\begin{threeparttable}
\begin{tabular}{c c c}
\hline
\toprule
\textbf {Layer} & \textbf{Network} & \textbf{Parameters} \\
\hline
\multirow{2}*{1} & Convolution layer& Filter=18, Kernel=2, Strides=1, Activation=Relu\\
\cline{2-3}
 & Max-pooling layer& Pool size=2, Strides=2\\
\hline
\multirow{2}*{2} & Convolution layer & Filter=18, Kernel=2, Strides=1, Activation=Relu \\
\cline{2-3}
 & Max-pooling layer & Pool size=2, Strides=2\\
\hline
\multirow{2}*{3} & Convolution layer & Filter=18, Kernel=2, Strides=1, Activation=Relu \\
\cline{2-3}
 & Max-pooling layer & Pool size=2, Strides = 2\\
\hline
4 & Fully connected layer & Unit =100, Activation =Relu, Dropout=0.2 \\
\hline
5 & LSTM layer & Cell=100, Dropout=0.2\\
\hline
6 & Fully connected layer & Unit=1, Activation=Sigmoid\\
\bottomrule
\end{tabular}
\end{threeparttable}
\end{center}
\end{table}

\begin{table}[t]
\scriptsize
\setlength{\abovecaptionskip}{0.cm}
\setlength{\belowcaptionskip}{-1cm}
\renewcommand{\arraystretch}{1.2}
\caption{Specific Configuration of Discriminator in proposed CR-GAN}
\label{Table3}
\begin{center}
\begin{threeparttable}
\begin{tabular}{c c c}
\hline
\toprule  
\textbf{Layer} & \textbf{Network} & \textbf{Parameters} \\
\hline
1 & LSTM layer & Cell=100, Dropout=0.2\\
\hline
2 & Fully connected layer & Unit=1, Activation=Sigmoid\\
\bottomrule
\end{tabular}
\end{threeparttable}
\end{center}
\end{table}

\begin{table}[!ht]
\scriptsize
\setlength{\abovecaptionskip}{0.cm}
\setlength{\belowcaptionskip}{-1cm}
\renewcommand{\arraystretch}{1.3}
\label{Table4}
\centering
\caption{Comparison of time complexity between the proposed method CR-GAN with its counterparts.}
\label{table_example}
\begin{center}
\begin{threeparttable}
\begin{tabular}{c c}
\toprule  
\multirow{1}*{\textbf{Method}} & \multicolumn{1}{c}{\textbf{Time complexity}}\\
\hline
RCGAN & $O(2W)$\\
\hline
T-CGAN & $O(\sum_{l=1}^{4} (s_l^2 n_l m_l^2))$\\
\hline
CR-GAN (Proposed) & $O(3\sum_{l=1}^{3} (s_l^2 n_l m_l^2)+4W)$\\
\bottomrule
\end{tabular}
\end{threeparttable}
\end{center}
\end{table}

\subsubsection{Multivariate time-series generation}
To verify the proposed method quantitatively, a ``Train on Synthetic data, Test on Real data'' provided in \cite{Ref18} is employed for quantitative comparison. Specifically, we use the synthetic data generated by CR-GAN to train a classifier model and then test it on a set of real examples. Besides, two competitive algorithms, RCGAN \cite{Ref18} and T-CGAN \cite{Ref17}, are employed for comparison. Considering only seven variables are used for time-series generation in RCGAN and T-CGAN, the same variables are employed as model input for a fair comparison. The employed variables in FD001 and FD003 can be observed from Fig. 6(a) and Fig. 6(b), respectively. The specific configurations of the proposed CR-GAN are listed in Tables II and III.

Fig. 6(a) summarizes all methods for FD001 using a classification evaluation index $AUROC$, which signifies how much a model could distinguish different classes. By checking the generated time-series of each variable, it is observed that the value of $AUROC$ for CR-GAN is very close to 1, indicating a high-quality generation ability. For RCGAN, it achieves the best result (95$\%$) for variable $S_{21}$, however, it is still worse than that of the proposed method (99$\%$). Here, $S_{21}$ stands for the twenty-first variable, and the subscript is used in the same way for other variables. Besides, the result of Variable $S_7$ in RCGAN is less than $60 \%$, yielding a poor accuracy. Similar results are observed for T-CGAN, which is also much worse than the proposed method. The corresponding results of FD003 are summarized in Fig. 6(b), and almost all values concerning index $AUROC$ are higher than 90$\%$, illustrating that the proposed model performs well.

\begin{table}[!ht]
\setlength{\abovecaptionskip}{0.cm}
\setlength{\belowcaptionskip}{-1cm}
\renewcommand{\arraystretch}{1.2}
\scriptsize
\caption {Comparisons between current RUL methods and their enhanced methods for engine data.}
\label{Table5}
\begin{center}
\begin{threeparttable}
\begin{tabular}{c c c c}
\hline
\toprule
\multirow{2}*{\textbf{Dataset}} & \multirow{2}*{\textbf{Approach}} & \multicolumn{2}{c}{\textbf{Indices}}\\
\cline{3-4}
& & $SF$ & $RMSE$ \\
\hline
\multirow{9}*{FD001} & TCMN & 1220.00 & 23.57\\
\cline{3-4}
& {TCMN with CR-GAN{$^\star$}} & 1027.00 & 18.39\\
\cline{3-4}
& Accuracy improvement ($\%$) & \textbf{15.82} & \textbf{21.98} \\
\cline{2-4}
& LSTM & 338.00 & 16.14 \\ 
\cline{3-4}
& {LSTM with CR-GAN{$^\star$}} & 301.06 & 15.30\\
\cline{3-4}
& Accuracy improvement ($\%$) & \textbf{10.93} & \textbf{5.21}\\
\cline{2-4}
& CapsNet & 276.34 & 12.58 \\ 
\cline{3-4}
& {CapsNet with CR-GAN{$^\star$}} & 216.17 & 12.49\\
\cline{3-4}
& Accuracy improvement ($\%$) & \textbf{21.77} & \textbf{0.72}\\
\cline{1-4}
\multirow{9}*{FD003} & TCMN & 1300.00 & 21.70\\
\cline{3-4}
& {TCMN with CR-GAN{$^\star$}} & 1257.00 & 17.92\\
\cline{3-4}
& Accuracy improvement ($\%$) & \textbf{3.31} & \textbf{17.42}\\
\cline{2-4}
& LSTM & 852.00 & 16.18 \\
\cline{3-4}
& {LSTM with CR-GAN{$^\star$}} & 257.32 & 15.58 \\
\cline{3-4}
& Accuracy improvement ($\%$) & \textbf{69.80} & \textbf{3.71}\\
\cline{2-4}
& CapsNet & 283.81 & 11.71 \\ 
\cline{3-4}
& {CapsNet with CR-GAN{$^\star$}} & 191.10 & 12.06\\
\cline{3-4}
& Accuracy improvement ($\%$) & \textbf{32.67} & \textbf{-2.98}\\
\bottomrule
\end{tabular}
    \begin{tablenotes}
     \item[$\star$] denotes the proposed methods.
   \end{tablenotes}
\end{threeparttable}
\end{center}
\end{table}

\begin{figure}[htb]
\subfigure[]
{
\begin{minipage}[t]{0.45\linewidth}
\centering
\includegraphics[width=4.5cm]{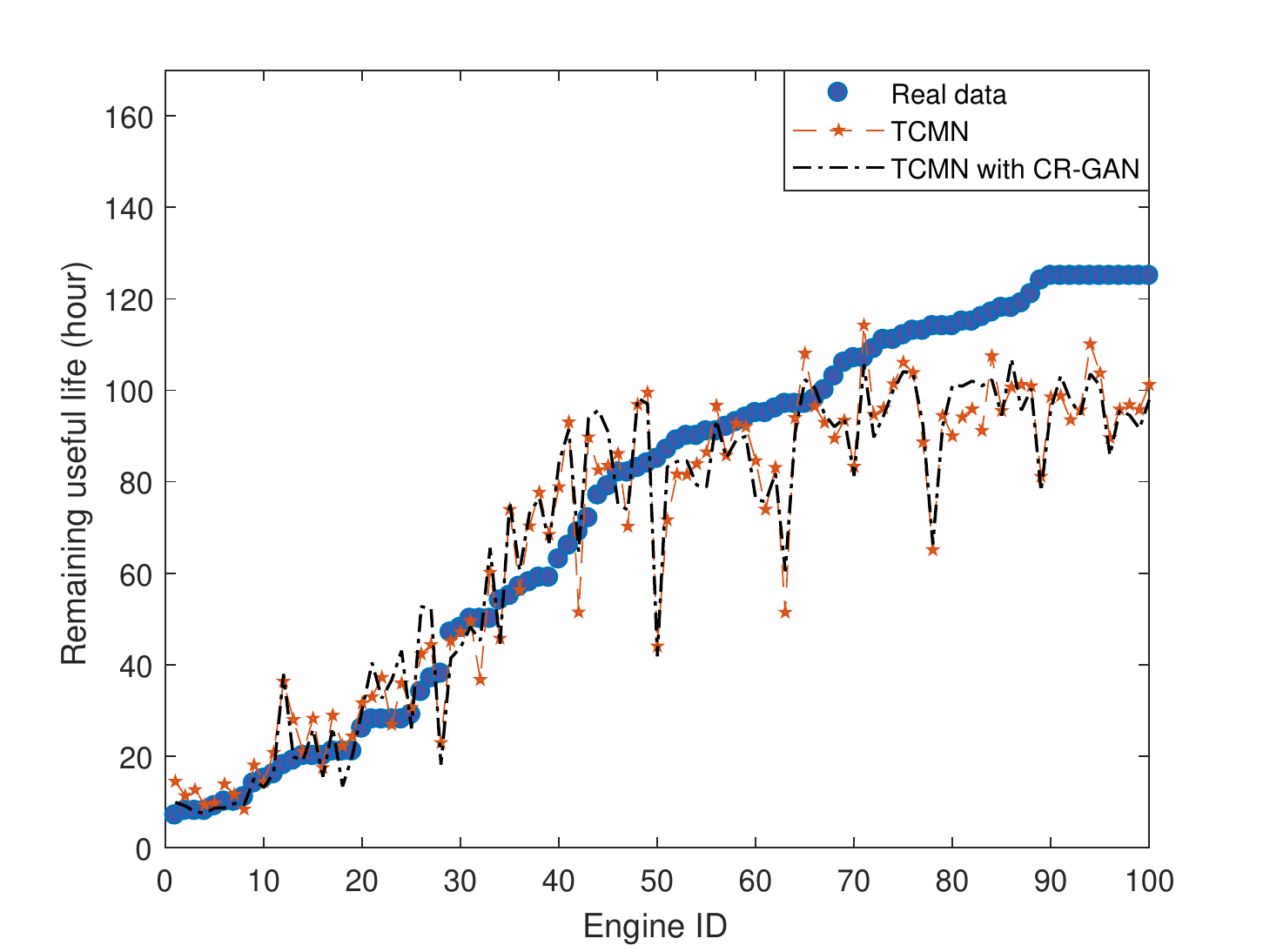}
\end{minipage}
}
\subfigure[]
{
\begin{minipage}[t]{0.45\linewidth}
\centering
\includegraphics[width=4.5cm]{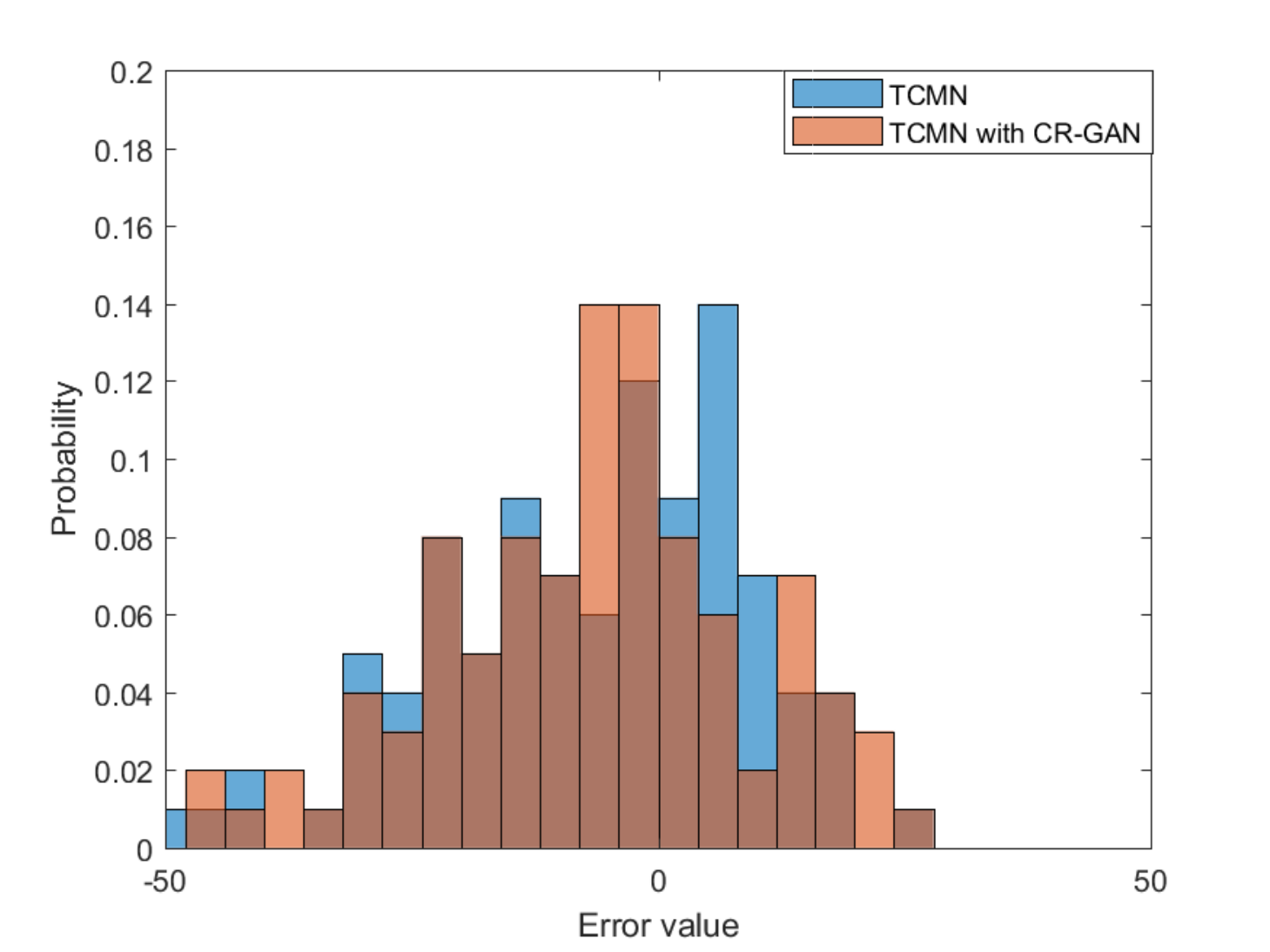}
\end{minipage}
}
\hfill
\subfigure[]
{
\begin{minipage}[t]{0.45\linewidth}
\centering
\includegraphics[width=4.5cm]{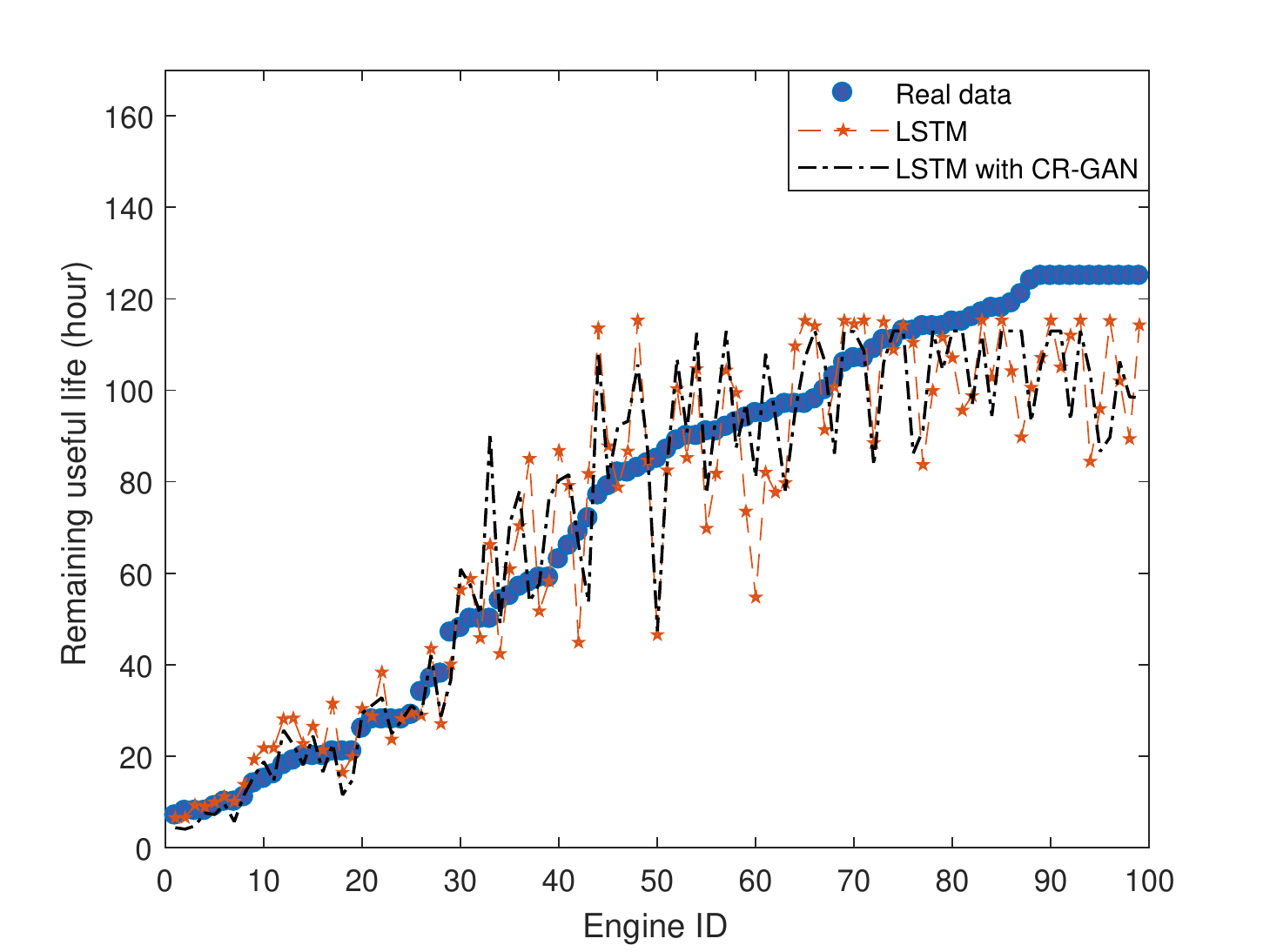}
\end{minipage}
}
\subfigure[]
{
\begin{minipage}[t]{0.45\linewidth}
\centering
\includegraphics[width=4.5cm]{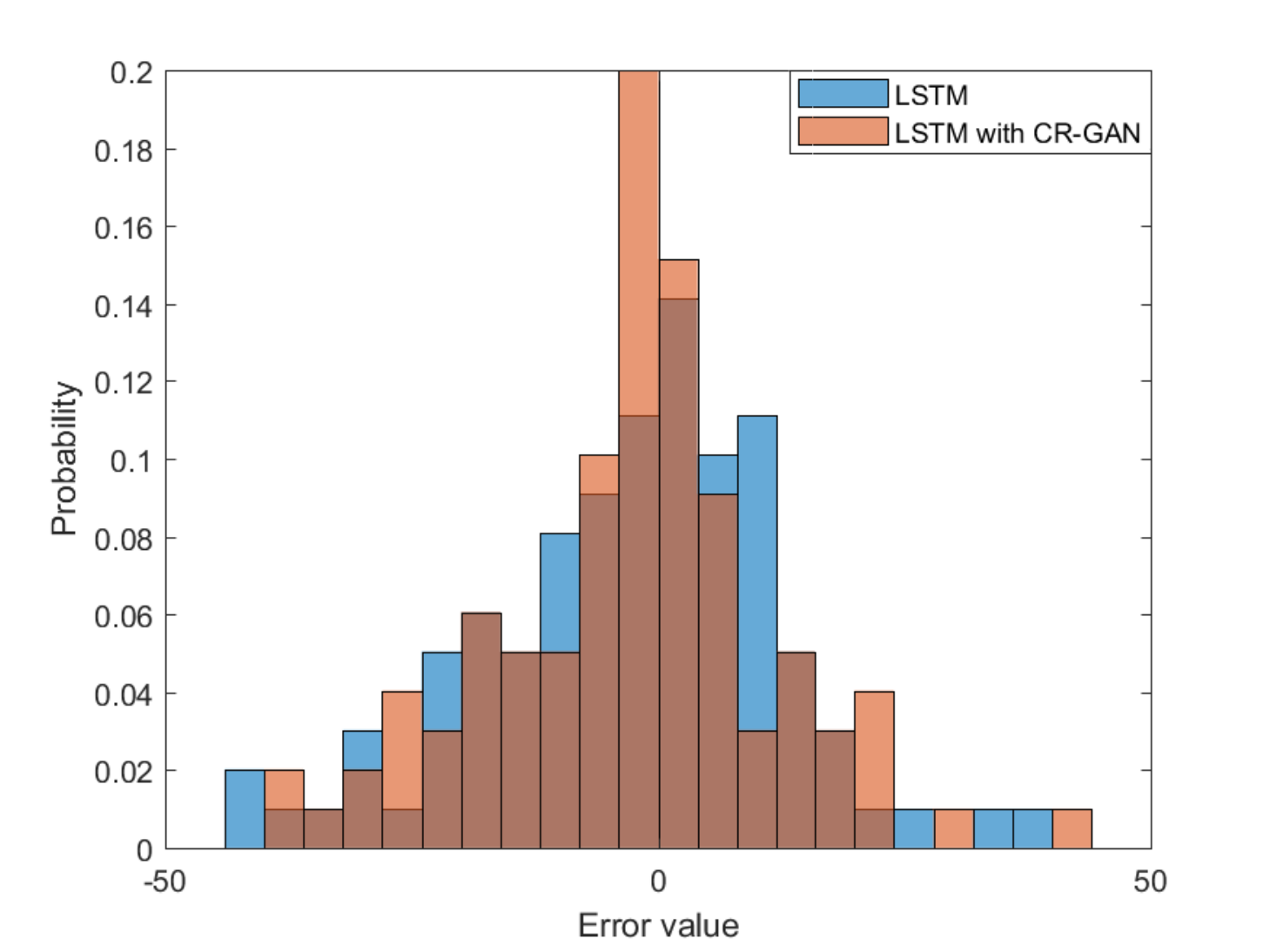}
\end{minipage}
}
\hfill
\subfigure[]
{
\begin{minipage}[t]{0.45\linewidth}
\centering
\includegraphics[width=4.5cm]{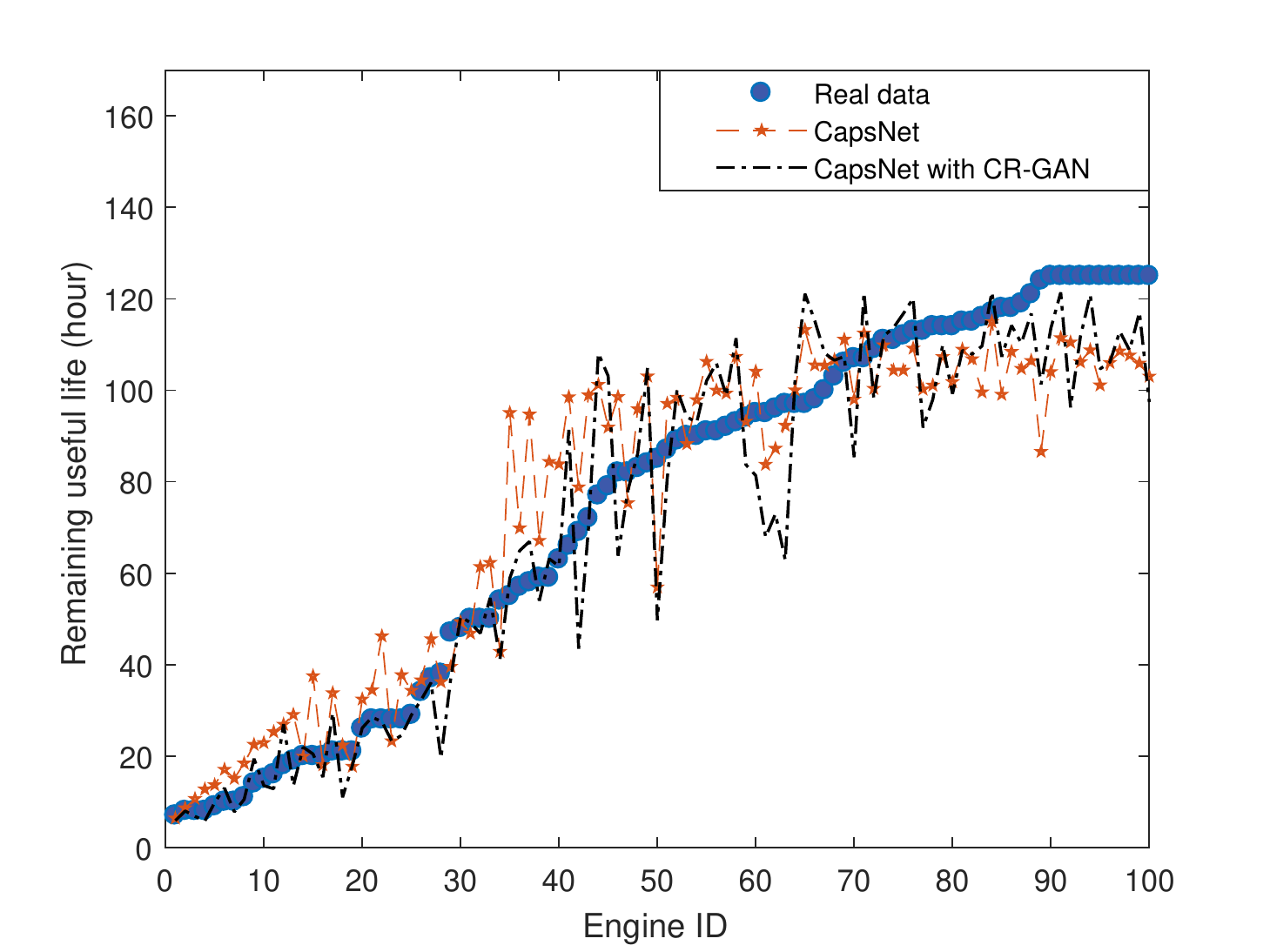}
\end{minipage}
}
\subfigure[]
{
\begin{minipage}[t]{0.45\linewidth}
\centering
\includegraphics[width=4.5cm]{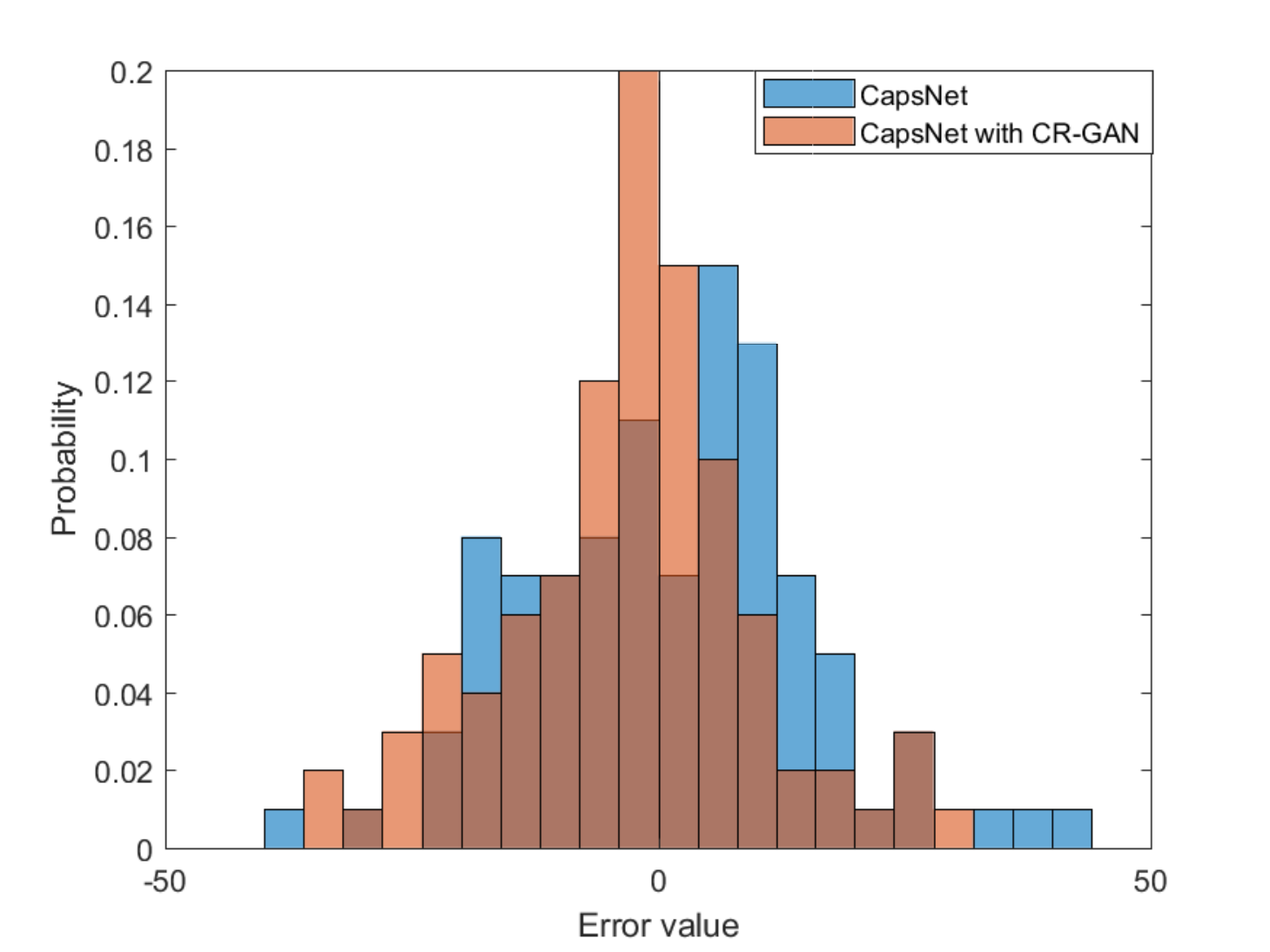}
\end{minipage}
}
\centering
\caption{Comparison on dataset FD001 for (a) Estimation error of TCMN and TCMN with CR-GAN, (b) Error distribution of TCMN and TCMN with CR-GAN, (c) Estimation error of LSTM and LSTM with CR-GAN, (d) Error distribution of LSTM and LSTM with CR-GAN, (e) Estimation error of CapsNet and CapsNet with CR-GAN, and (f) Error distribution of CapsNet and CapsNet with CR-GAN.}
\label{MyFig7}
\end{figure}

With the network architecture presented in Tables II and III, the overall time complexity of the proposed method is calculated by a linear combination of the generator and the discriminator. Specifically, when the input channel is one, the time complexity of the $l^{th}$ convolutional layer is estimated as $O(s_l^2 n_l m_l^2)$ \cite{Ref34}, where $n_l$ is the number of filters in the $l^{th}$ layer (18 here), $s_l$ is the spatial size of the filter (2 here), and $m_l$ is the spatial size of the output feature map. Since the RNN/LSTM is local in space and time, the complexity of an LSTM/RNN layer per time step is equal to $O(W)$, where $W$ is the number of weight. Finally, the time complexity of combining the generator and the discriminator is calculated as $O(3\sum_{l=1}^{3} (s_l^2 n_l m_l^2)+4W)$ through the linear superposition of all convolutional layers and LSTM layers. The time complexity of RC-GAN and T-CGAN can be computed in the same way. As shown in Table IV, it is observed that the time complexity of the proposed method is a litter higher than that of RC-GAN and T-CGAN since more CNN and LSTM layers are used. In comparison with RCGAN and T-CGAN, the increased training time of the proposed method is not significant. Moreover, once the offline model has been well trained, it could be directly adopted in the online estimation stage with rapid computing.

\subsubsection{Enhanced RUL estimation with generated time-series}
With the CR-GAN model, performances of the enhanced RUL estimation is verified by feeding the generated time-series into three typical RUL estimation methods \cite{Ref5}, \cite{Ref7}, \cite{Ref9} with model re-training. The same model structure listed in Tables II and III are adopted.

\begin{figure}[htb]
\subfigure[]
{
\begin{minipage}[t]{0.45\linewidth}
\centering
\includegraphics[width=4.5cm]{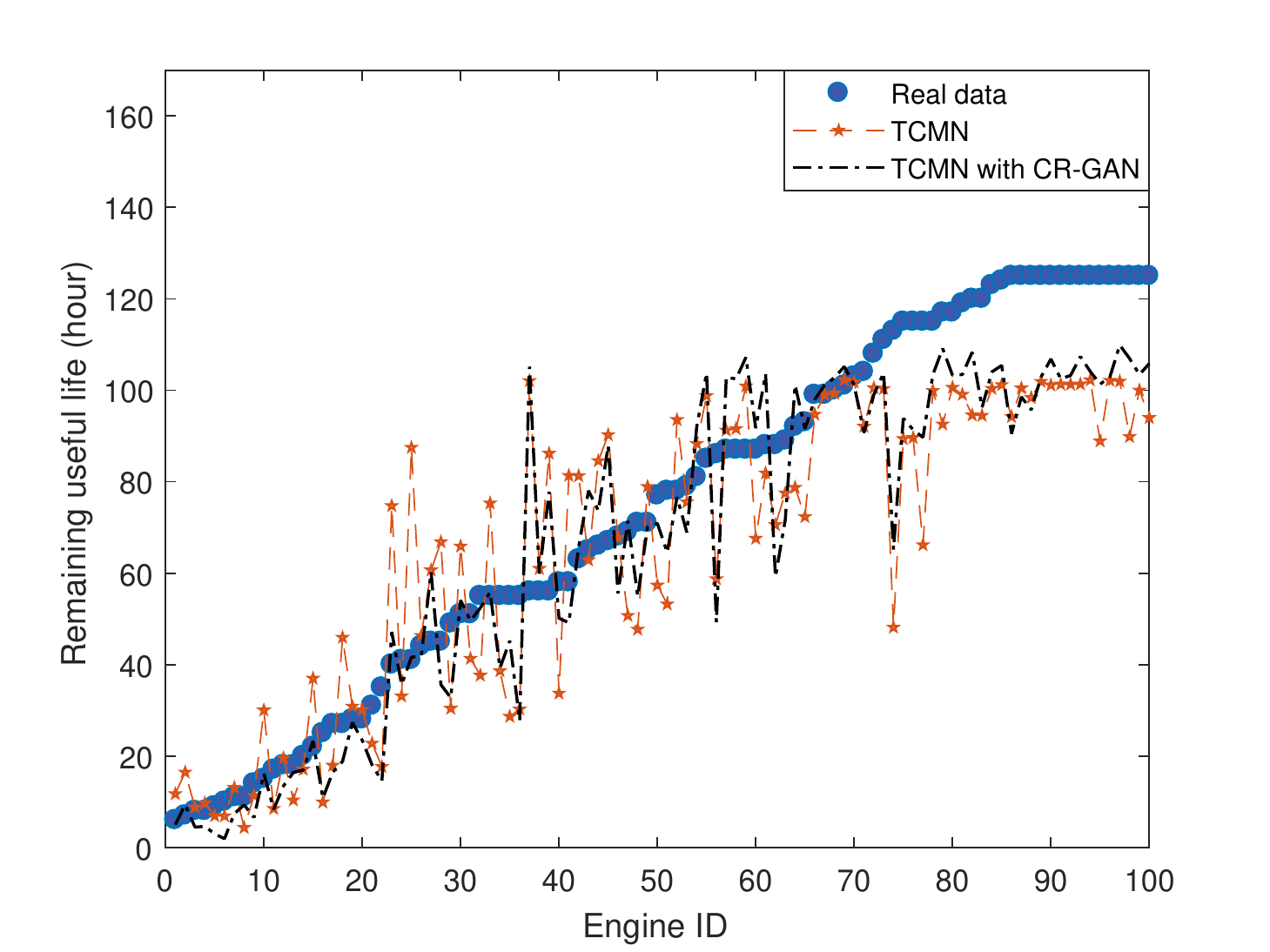}
\end{minipage}
}
\subfigure[]
{
\begin{minipage}[t]{0.45\linewidth}
\centering
\includegraphics[width=4.5cm]{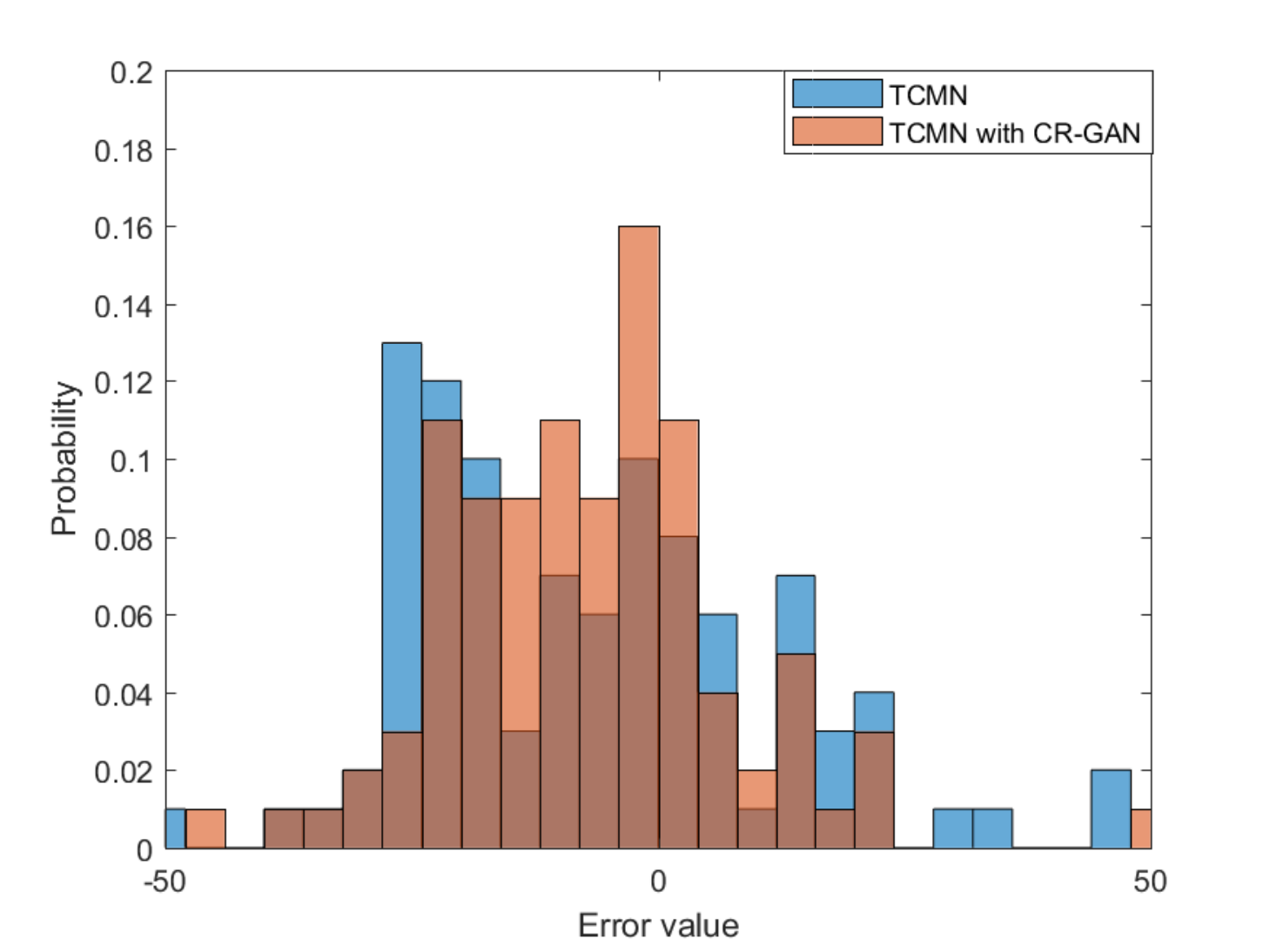}
\end{minipage}
}
\hfill
\subfigure[]
{
\begin{minipage}[t]{0.45\linewidth}
\centering
\includegraphics[width=4.5cm]{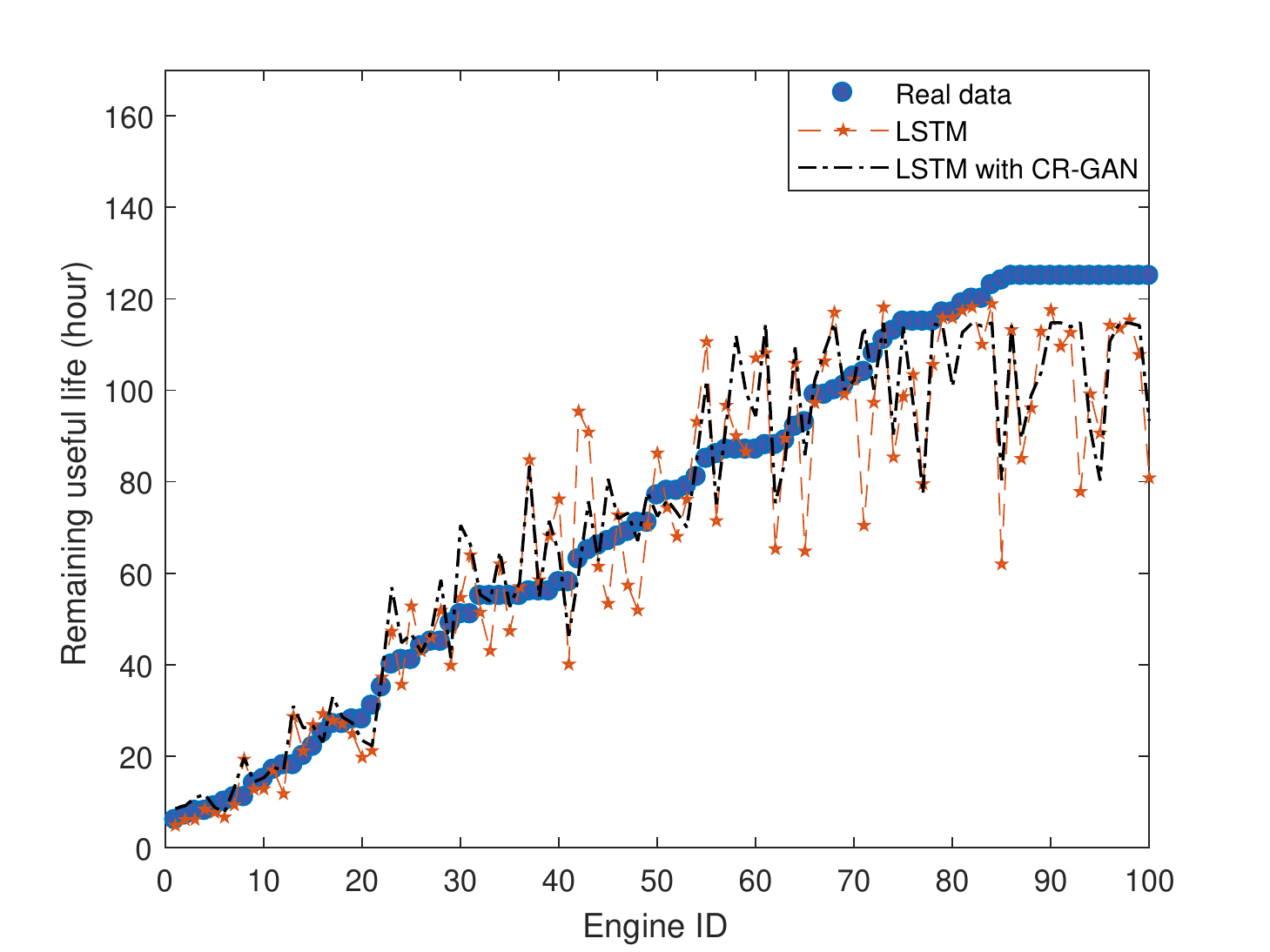}
\end{minipage}
}
\subfigure[]
{
\begin{minipage}[t]{0.45\linewidth}
\centering
\includegraphics[width=4.5cm]{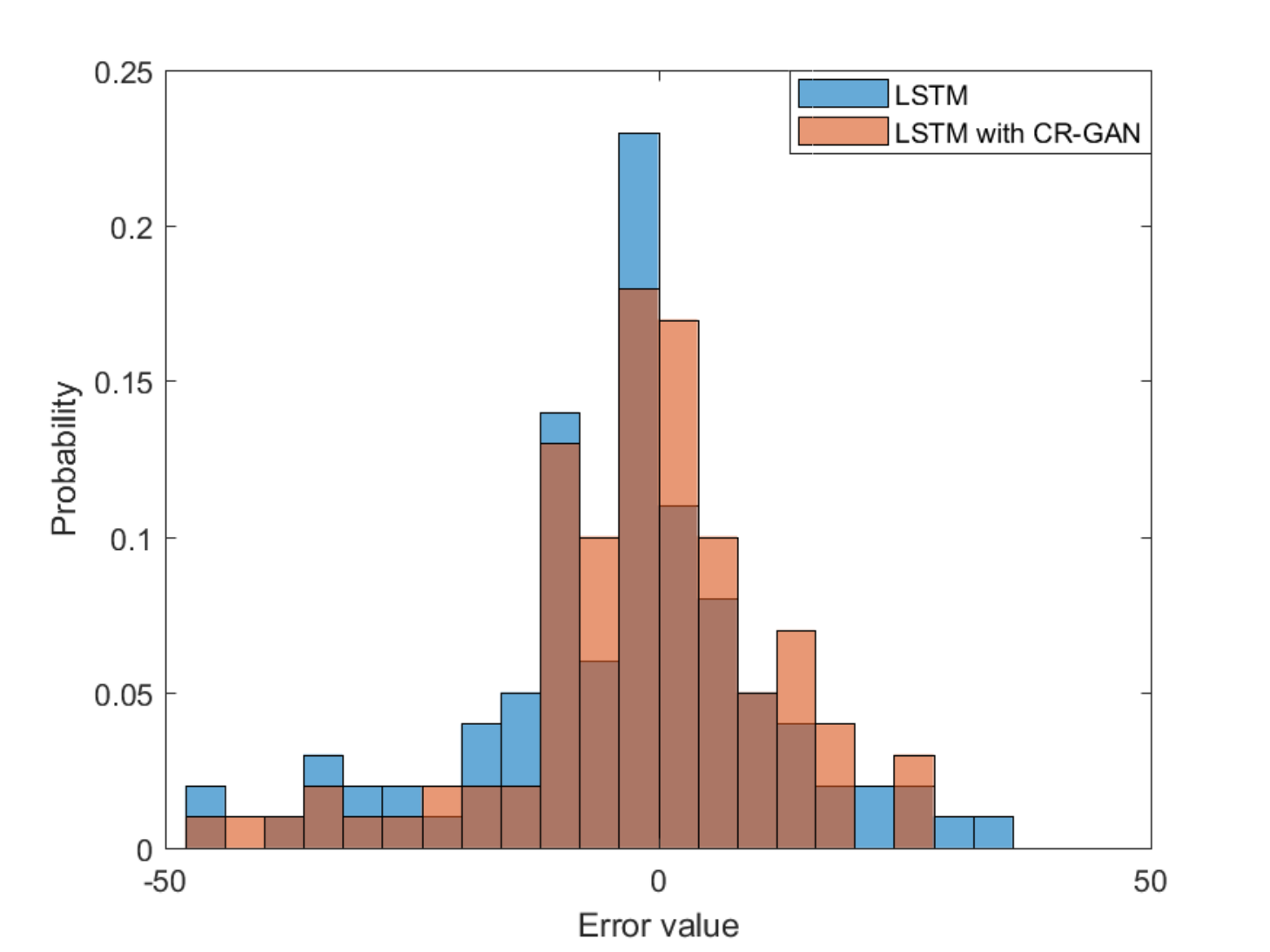}
\end{minipage}
}
\hfill
\subfigure[]
{
\begin{minipage}[t]{0.45\linewidth}
\centering
\includegraphics[width=4.5cm]{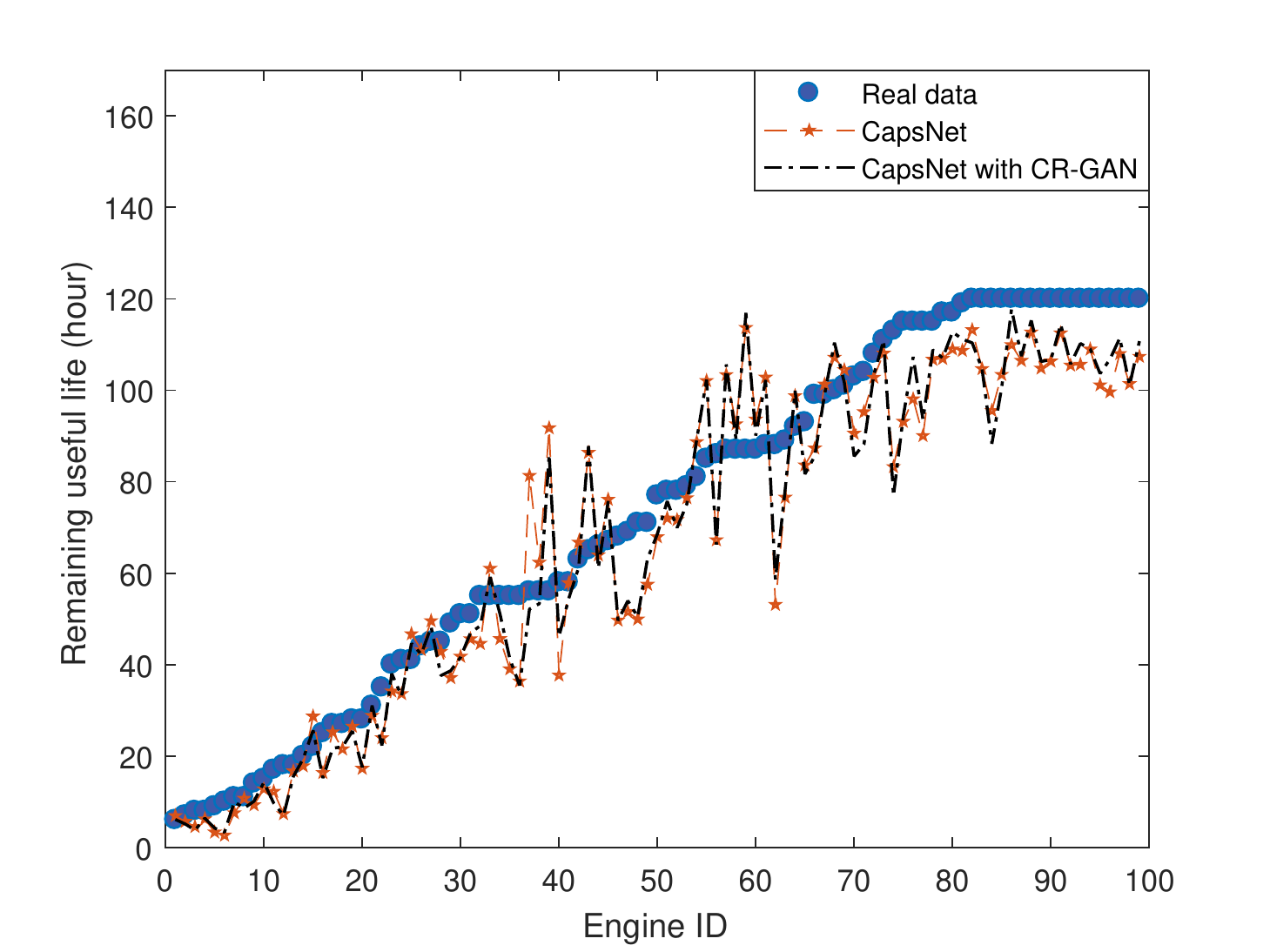}
\end{minipage}
}
\subfigure[]
{
\begin{minipage}[t]{0.45\linewidth}
\centering
\includegraphics[width=4.5cm]{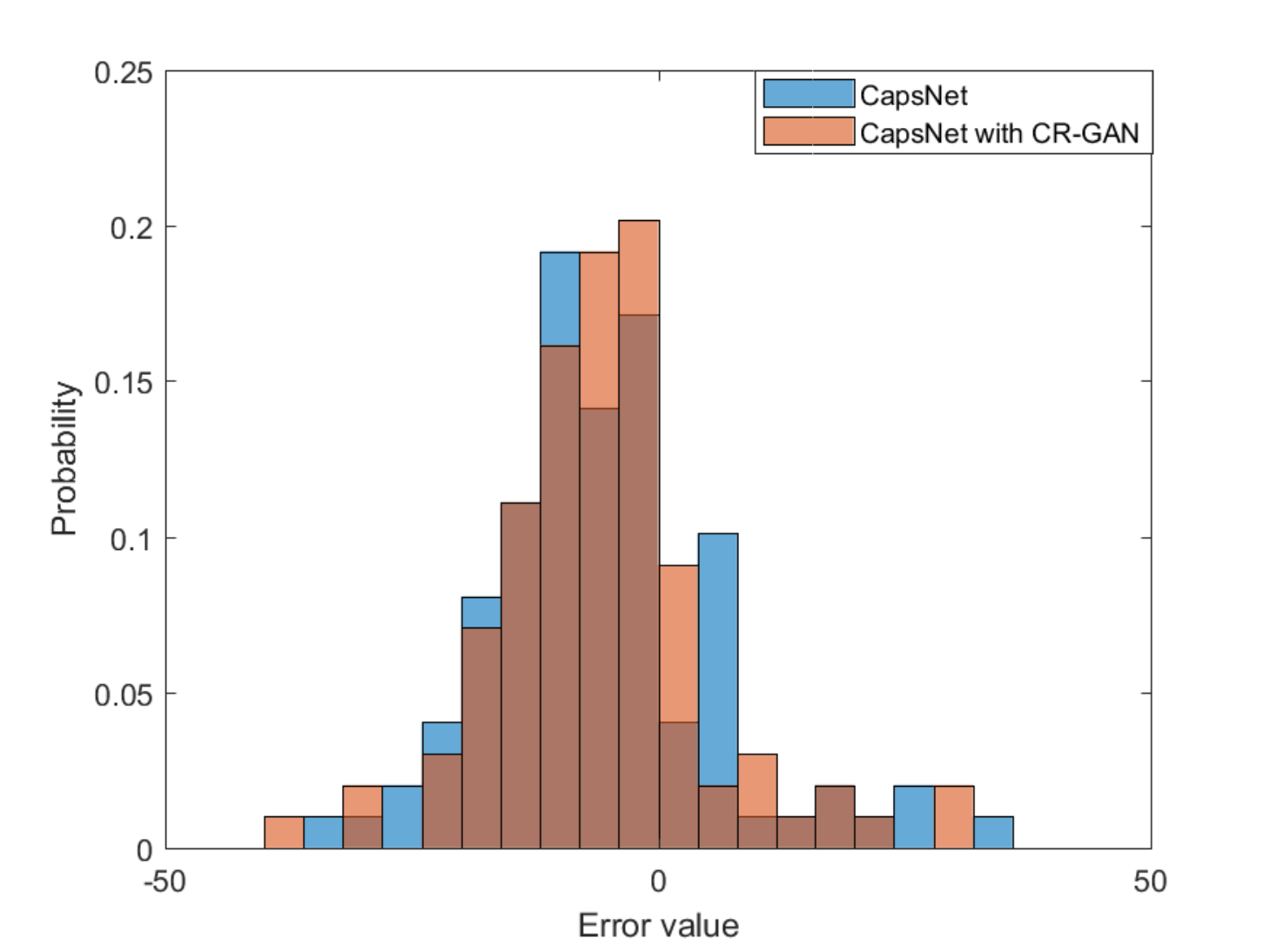}
\end{minipage}
}
\centering
\caption{Comparison on dataset FD003 for (a) Estimation error of TCMN and TCMN with CR-GAN, (b) Error distribution of TCMN and TCMN with CR-GAN, (c) Estimation error of LSTM and LSTM with CR-GAN, (d) Error distribution of LSTM and LSTM with CR-GAN, (e) Estimation error of CapsNet and CapsNet with CR-GAN, and (f) Error distribution of CapsNet and CapsNet with CR-GAN.}
\label{MyFig8}
\end{figure}

It should be noted that all variables are used for RUL estimation modeling. This is different from the purpose of data generation in the last subsection, which only considers fair comparison with existing methods. First, data preparation is arranged to remove redundant variables from both training and testing data. By checking mean and variance of each variable, variables with constant value are Variables $S_1, S_5, S_6, S_{10}, S_{16}, S_{18}, S_{19}$ in FD001 and Variables $S_1, S_5, S_{10}, S_{16}, S_{18}, S_{19}$ in FD003, which are non-informative. Next, the degradation of the engine system typically starts after a certain degree of usage, resulting in the less importance of the normal data samplings. According to apriori knowledge, the samplings in the last 125 hours in each cycle have remained for data generation, and the window size $L$ is chosen as 100 here. In this way, the number of original time-series is enlarged 26 times to enable basic training of CR-GAN. With data normalization in the way given in Eq. 6, the inputs of CR-GAN is normalized data matrixes with dimensions $2600 \times 14 \times \ 100$ and dimensions $2600 \times 15 \times \ 100$ for FD001 and FD003, respectively.

\begin{table*}[!htb]
\scriptsize
\setlength{\abovecaptionskip}{0.cm}
\setlength{\belowcaptionskip}{-1cm}
\renewcommand{\arraystretch}{1.2}
\caption{RUL estimation error comparison between the enhanced LSTM and original LSTM for cyclic degradation battery data.}
\label{Table6}
\begin{center}
\begin{threeparttable}
\begin{tabular}{c c c c c c c}
\hline
\textbf{Battery Name} & \textbf{Method} & \textbf{Number of Generated Time-series} & \textbf{Actual RUL} & \textbf{Predicted RUL} & \textbf{RUL Error} & \textbf{RMSE} \\
\hline
\multirow{4}*{B0005} & LSTM & -- & \multirow{4}*{74} & 94 & 20 & 0.113 \\
\cline{2-2}
& \multirow{3}*{LSTM with CR-GAN{$^\star$}} & 50 & & 78 & 4 & 0.0822\\
& & 100 & & 76 & 2  & 0.0814 \\
& & 150 & & 73 & -1 & 0.0563 \\
\hline
\multirow{4}*{B0006} & LSTM & -- & \multirow{4}*{58} & 106 & 48 & 0.1216 \\
\cline{2-2}
& \multirow{3}*{LSTM with CR-GAN{$^\star$}} & 50 & & 77 & 19 & 0.0852\\
& & 100 & & 73 & 15 & 0.0699 \\
& & 150 & & 57 & -1 & 0.0722 \\
\hline
\multirow{4}*{B0018} & LSTM & -- & \multirow{4}*{46} & 66 & 20 & 0.0607 \\
\cline{2-2}
& \multirow{3}*{LSTM with CR-GAN{$^\star$}} & 50 & & 46 & 0 & 0.0441\\
& & 100 & & 46 & 0 & 0.0343 \\
& & 150 & & 46 & 0 & 0.0239 \\
\bottomrule
\end{tabular}
\begin{tablenotes}
     \item[$\star$] denotes the proposed methods.
\end{tablenotes}
\end{threeparttable}
\end{center}
\end{table*}

\begin{figure*}[!htb]
\centering
\subfigure[Battery B0005]{
\begin{minipage}[t]{0.32\linewidth}
\centering
\includegraphics[width=6.6cm]{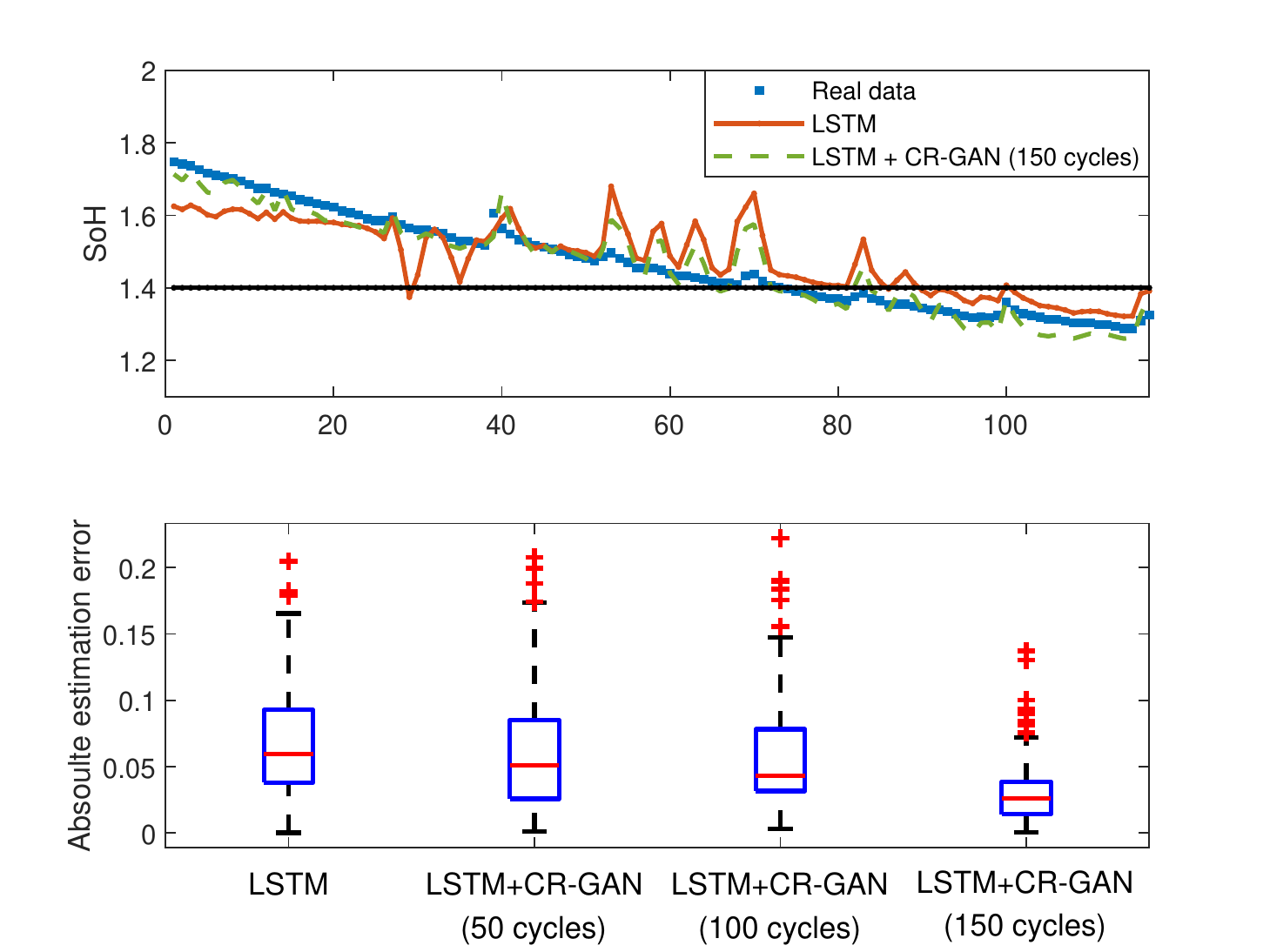}
\end{minipage}}
\subfigure[Battery B0006]{
\begin{minipage}[t]{0.32\linewidth}
\centering
\includegraphics[width=6.6cm]{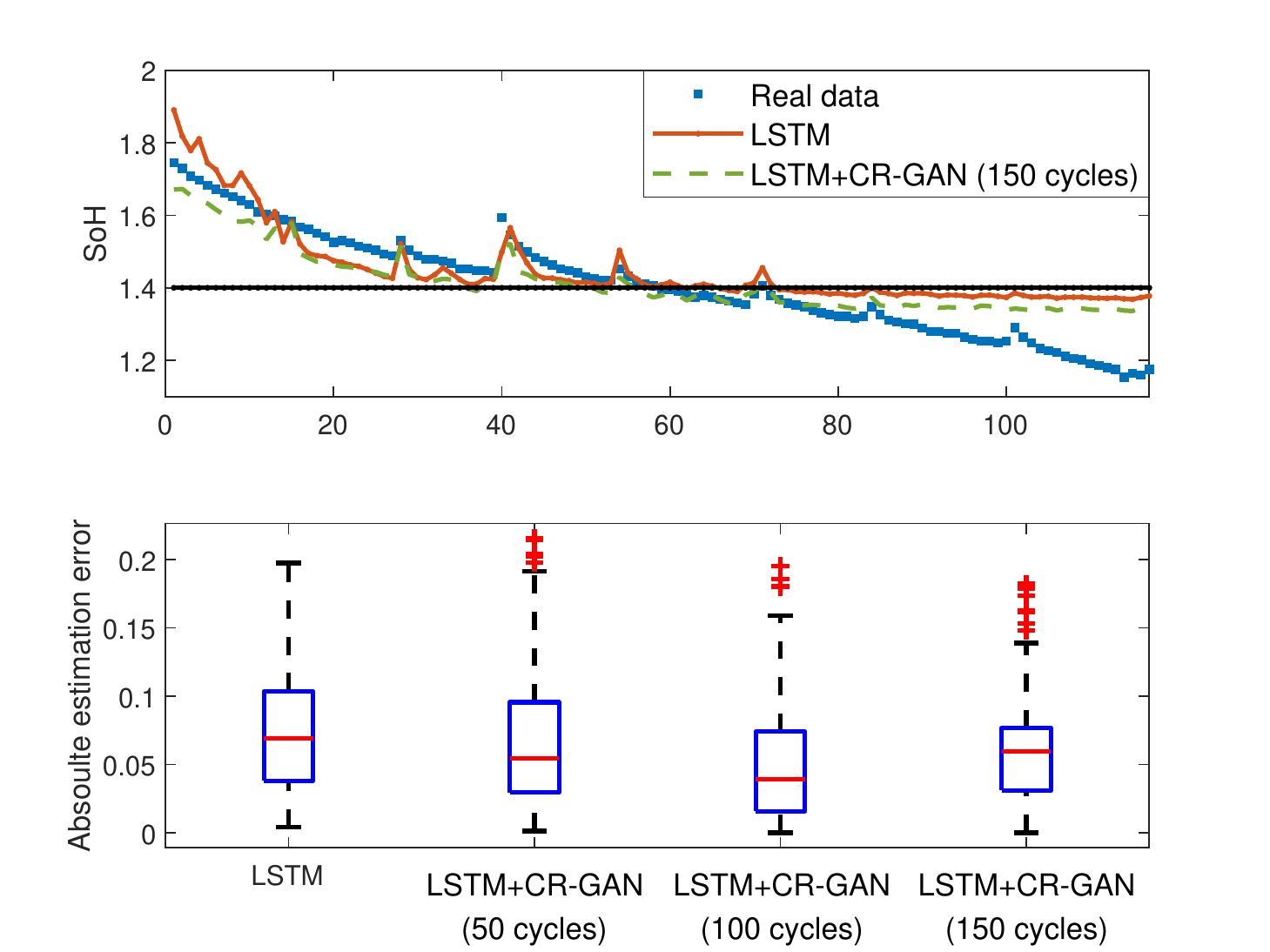}
\end{minipage}}
\subfigure[Battery B0018]{
\begin{minipage}[t]{0.32\linewidth}
\centering
\includegraphics[width=6.6cm]{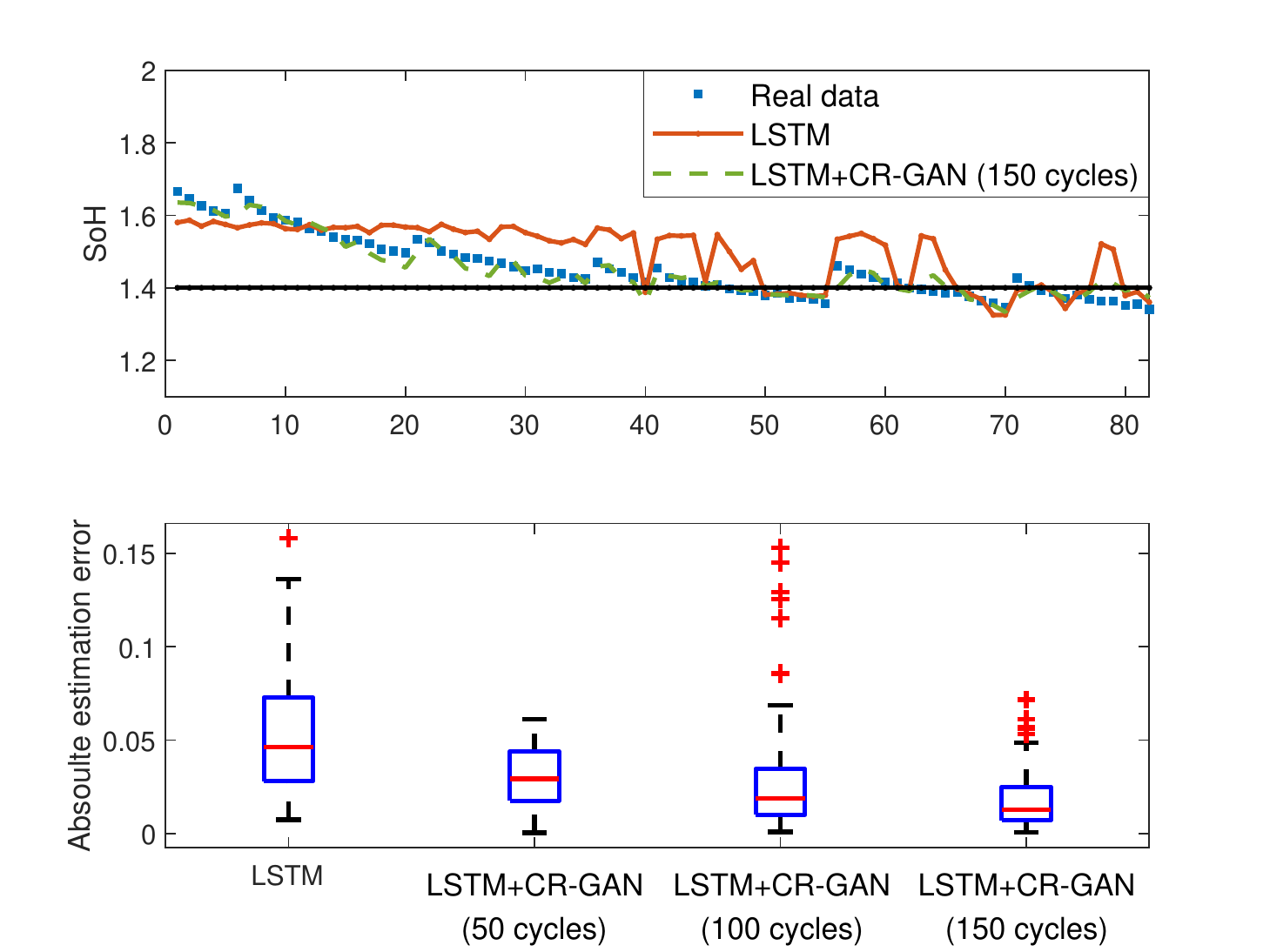}
\end{minipage}}
\centering
\caption{Accuracy comparison between LSTM and its enhanced framework for (a) Battery B0005, (b) Battery B0006, and (c) Battery B0018.}
\label{MyFig9}
\end{figure*}

After filtering out irrational time-series, a total number of 200 run-to-failure time-series with a length of 100 is generated using CR-GAN. These synthetic time-series are mixed with original time-series to form a new training dataset for the RUL estimation model. Here, three baseline RUL estimation methods, LSTM \cite{Ref5}, temporal convolution neural network (TCMN) \cite{Ref7}, and CapsNet \cite{Ref9}, are employed for comparison. With the given steps through Sections III. A to III. E, Table V summarizes the estimation results of all enhanced methods using two criteria, including $SF$ and $RMSE$. For the enhanced TCMN, the value of $RMSE$ is reduced from 23.57 to 18.39 for FD001 and from 21.70 to 17.92 for FD003, respectively. With respect to $SF$, the value is also significantly reduced using the proposed framework. For the enhanced LSTM algorithm, the estimation error of $RMSE$ is reduced from 16.14 to 15.30 for FD001 and from 16.18 to 15.58 for FD003, respectively. CapsNet \cite{Ref31} proposed by Hinton in 2017 aims at extracting vector-shaped features to model the hierarchical relationship. The state-of-art results of $RMSE$ index reported by CapsNet are 12.59 and 11.71  for FD001 and FD003, respectively \cite{Ref9}. The proposed framework helps CapsNet further reducing the value of the index $SF$ from 276.34 to 216.17 for FD001 and from 283.81 to 191.10 for FD003, respectively. More precisely, in Table V, improvements are calculated in the ratio that the reduced errors are divided by the values of its reference model.

As an auxiliary evaluation to $RMSE$ and $SF$, the estimation error of the specific engine and the corresponding distribution are shown in Figs. 7 and 8 for FD001 and FD003, respectively. The engine index here is sorted according to the ascending of real RUL. It provides an intuitive way to observe the performance by analyzing each engine. It is observed that adding generated time-series will drive the estimations closer to their ground truth. Further, the distribution of estimation errors is shown in the histogram for the first time to account for the percentage of error band with a specific interval, which is four here. For the similar values of $RMSE$ and $SF$, a method is regarded better if more estimation errors are distributed in the low-value zones. In Figs. 7(b), 7(d), and 7(f), it is observed that error distributions of the proposed method for a higher ratio in low-value zones in comparison with their counterparts. A similar conclusion can be drawn for FD003 through analyzing Figs. 8(b), 8(d), and 8(f).

\subsection{RUL Estimation for Cyclic Degradation Pattern}
As a reusable component, LiB suffers from performance degradation due to inherent electrochemical reactions during repetitive usage. A LiB needs to be replaced when its real capacity drops to a certain threshold, i.e., EOL, which is usually determined as 70$\%$ of its rated capacity \cite{Ref32}. Correspondingly, RUL is calculated as the difference between predicted EOL and the real value. In practice, discharging current and discharging voltage could be measured for capacity inference for RUL estimation. Having the data characteristics of cycling run-to-failure time-series, LiB is an ideal benchmark to test the performance of the proposed method concerning cyclic degradation pattern. Here, LiB data provided by NASA Ames Prognostics Center of Excellence \cite{Ref23} are employed for testing the performance of the enhanced RUL framework.

\subsubsection{Multivariate time-series generation}
Datasets B0005, B0006, and B0018 are selected for analysis, which have been used in \cite{Ref33} that adopts LSTM for RUL estimation. For cycling data in each dataset, current signal, voltage signal, and ground truth capacity are employed as the generated variables, and the time index is used as the condition information. In \cite{Ref33}, two datasets and the first 50 cycles in the third one are used for training in turn, and the remaining ones are used for testing. According to steps given in Section III, 150 cycling data located in the first 50 cycles are generated. Here, the same configurations listed in Tables II and III are adopted with $L=200$.

\subsubsection{Enhanced RUL estimation with generated time-series}
Employing the same LSTM model structure in \cite{Ref33}, the RUL estimation ability of the proposed framework is further tested. Taking current and voltage signals as input variables, the estimation model outputs the prediction of capacity. Moreover, to reveal the influences of data amount on RUL estimation accuracy, three different amounts of generated time-series are considered, including 50, 100, and 150. According to Table VI, two conclusions can be clearly drawn. First, both $RMSE$ and RUL errors have been significantly reduced with the introduction of generated time-series. Especially, RUL estimation error has been reduced to zero for Battery B0018 when the number of generated time-series is 150 cycles. Second, the more generated data used for model re-training, the smaller RUL error will be, which can be observed from the value of index $RMSE$. To intuitively show the estimation results, Fig. 9 compares LSTM and its enhanced model when 150 generated cycles are used for model training. The y-label of the top subplot of Fig. 9 presents the state of health (SoH), which indicates the health status of a battery by informing the remaining capacity or point that has been reached in the life cycle. Besides, the absolute error distribution of each model with respect to SoH is shown in Fig. 9 using the boxplot.

From the above analysis for both types of degradation systems, RUL estimation errors have significantly been reduced with the aid of regeneration of run-to-failure time-series, demonstrating the efficacy and success of the proposed framework.

\section{Conclusion and Future work}
To effectively address the challenge of data shortage of run-to-failure time-series, this article puts forward the idea of time-series generation for the first time in improving RUL estimation. It is a self-learning enhancement framework aiming at integrating time-series generation into current RUL estimation models. The specificity of this article is to synthesize realistic multivariate time-series with a novel network named convolutional recurrent conditional generative adversarial network (CR-GAN), yielding a wealth of data to escalate the training procedure of existing RUL models. Moreover, the degradation nature of various systems is well studied, producing the classification of non-cyclic and cyclic degradation patterns. By comparing the current RUL estimation models with their enhanced versions, practical experiment results on both aero-engine system, and Lithium-ion aging clearly demonstrate the strengths of the proposed framework in terms of diverse evaluation indices. Therefore, it would be interesting to apply this idea for more industrial scenarios where it is economically expensive or dangerous to gain enough run-to-failure data.

This article verifies the feasibility of employing CR-GAN for generating time-series for better RUL estimation. However, there are still some possible directions deserving more efforts:

1) Run-to-failure time-series possess plenty of statistical properties, such as slow-varying trends, etc. It is possible that the objective function in the original GAN is not able to preserve all the statistics of the real time-series. It may be promising to consider proper statistical constraints in the objective function of the original GAN to obtain more reliable time-series.

2) Gaussian distribution has been widely selected to present the given input noise vector in the image processing field. Although Gaussian distribution is continually adopted for time-series generation, it is interesting to investigate the performance when other noise statistics are concerned.

3) Although irrational generated time-series could be filtered out using a post-processing way, it is recommended to consider a direct way during adversarial learning. Especially, irrational time-series may be easily recognized by experienced engineers equipped with plentiful domain knowledge. Therefore, it may provide a feasible solution to filter out irrational time-series with a discriminator considering domain knowledge.

\ifCLASSOPTIONcaptionsoff
  \newpage
\fi

\end{document}